\pdfoutput=1 

\documentclass[runningheads]{llncs}
\usepackage{graphicx}
\usepackage{amsmath,amssymb} 
\usepackage{color}

\usepackage[table]{xcolor}
\usepackage{soul}
\usepackage{booktabs}
\usepackage[normalem]{ulem}

\usepackage{varwidth}
\usepackage[font=small,labelfont=bf]{caption}

\let\subparagraph\relax

\newif\ifdraft
\draftfalse

\definecolor{orange}{rgb}{1,0.5,0}
\definecolor{violet}{RGB}{70,0,170}
\definecolor{magenta}{RGB}{170,0,170}
\definecolor{dgreen}{RGB}{0,150,0}

\usepackage[normalem]{ulem}

\ifdraft
 \newcommand{\PF}[1]{{\color{red}{\bf PF: #1}}}
 
 \newcommand{\WL}[1]{{\color{blue}{\bf WL: #1}}}
 
  \newcommand{\MS}[1]{{\color{dgreen}{\bf MS: #1}}}
 
\else
 \newcommand{\PF}[1]{}
 
 \newcommand{\WL}[1]{}
 
 \newcommand{\MS}[1]{}
  
\fi

\newcommand{\comment}[1]{}
\newcommand{\parag}[1]{\vspace{-3mm}\paragraph{#1}}
\newcommand{\sparag}[1]{\subparagraph{#1}}

\newcommand{\bA}{\mathbf{A}}

\newcommand{\bD}{\mathbf{D}}
\newcommand{\bI}{\mathbf{I}}

\newcommand{\bP}{\mathbf{P}}

\newcommand{\bZ}{\mathbf{Z}}

\newcommand{\z}{\mathbf{z}}

\newcommand{\fd}{\mathcal{F}_{d}}
\newcommand{\fz}{\mathcal{F}_{z}}

\newcommand{\FGSMU}[1]{\textbf{FGSM-U(#1)}}
\newcommand{\FGSMT}[1]{\textbf{FGSM-T(#1)}}
\newcommand{\FGSMUE}[1]{\textbf{FGSM-UE(#1)}}
\newcommand{\FGSMTE}[1]{\textbf{FGSM-TE(#1)}}

\usepackage{multirow}

\begin{document}
\pagestyle{headings}
\mainmatter

\title{Using Depth for Pixel-Wise Detection of Adversarial Attacks in Crowd Counting} 
\titlerunning{Pixel-Wise Adversarial Detection in Crowd Counting}

\author{
	Weizhe Liu\textsuperscript{}
	\quad
	Mathieu Salzmann\textsuperscript{}
	\quad
	Pascal Fua\textsuperscript{}\\
}
  \authorrunning{W. Liu et al.}
  %
  \institute{Computer Vision Laboratory, \'{E}cole Polytechnique F\'{e}d\'{e}rale de Lausanne (EPFL)
  \email{\{weizhe.liu, mathieu.salzmann, pascal.fua\}@epfl.ch}}

\maketitle



\begin{abstract}

State-of-the-art methods for counting people in crowded sce\-nes rely on deep networks to estimate crowd density. While effective, deep learning approaches are vulnerable to adversarial attacks, which, in a crowd-counting context, can lead to serious security issues. However, attack and defense mechanisms have been virtually unexplored in regression tasks, let alone for crowd density estimation.

In this paper, we investigate the effectiveness of existing attack strategies on crowd-counting networks, and introduce a simple yet effective pixel-wise detection mechanism. It builds on the intuition that, when attacking a multitask network, in our case estimating crowd density and scene depth, both outputs will be perturbed, and thus the second one can be used for detection purposes. We will demonstrate that this significantly outperforms heuristic and uncertainty-based strategies.

\end{abstract}


\section{Introduction}

State-of-the-art crowd counting algorithms~\cite{Zhang15c,Zhang16s,Onoro16,Sam17,Xiong17,Sindagi17,Shen18,Liu18b,Li18f,Sam18,Shi18,Liu18c,Idrees18,Ranjan18,Cao18} rely on Deep Networks to regress a crowd density, which is then integrated to estimate the number of people in the image. Their application can have important societal consequences, for example when they are used to assess how many people attended a demonstration or a political event. 

In the ``Fake News" era, it is therefore to be feared that hackers might launch adversarial attacks to bias the output of these models for political gain. While such attacks have been well studied for classification networks~\cite{Goodfellow2014a,Kurakin16,Madry18,Moosavi16,Moosavi17,Carlini17}, they remain largely unexplored territory for people counting and even for regression at large. The only related approach we know of~\cite{Ranjan19a} is very recent and specific to attacking optical flow networks, leaving the pixel-wise detection of attacks untouched.  

In this paper, our goal is to blaze a trail in that direction. Our main insight is that if a two-stream network is trained to regress both the people density and the scene depth, it becomes very difficult to affect one without affecting the other. In other words, pixels that have been modified to alter the density estimate will also produce incorrect depths, which can be detected by estimating depth using unrelated means. As we will show, this can be done using a depth sensor, simple knowledge about the scene geometry, or even an unrelated deep model pre-trained to estimate depth from images. In other words, our approach works best for a static camera for which the depth of the scene it surveys can be accurately computed but remains applicable to mobile cameras that make such computation more difficult. This assumes that such reference depth maps can be kept safe from the attacker. When this cannot be done, we will show that using statistics from depth maps acquired earlier suffices to detect tampering at a later date. 
 
\begin{figure}[t]
    \centering
    \includegraphics[width=1.0\linewidth]{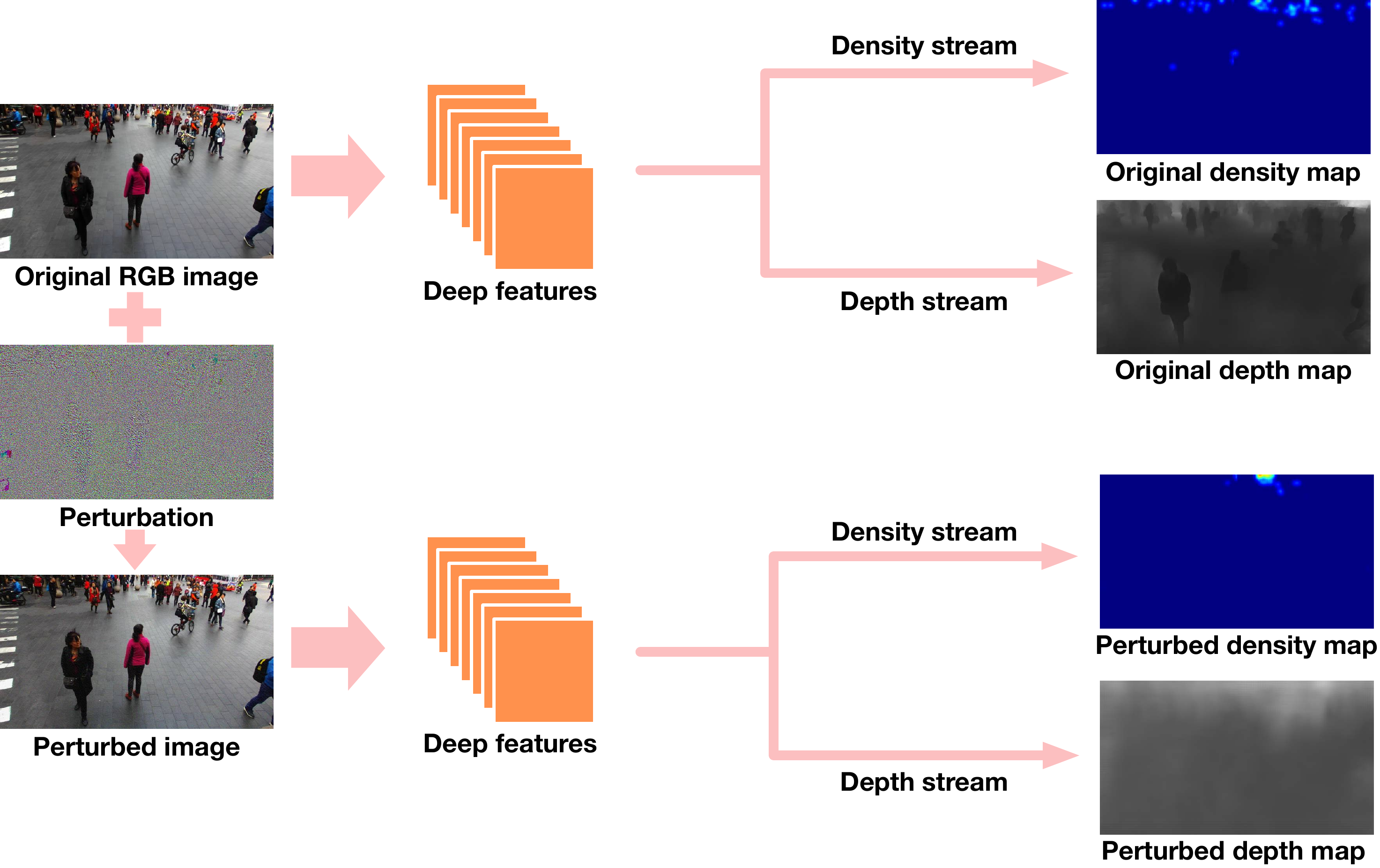}
    \vspace{-5mm}
    \caption{ {\bf Density and Depth (DaD) model.} A two-stream network is trained to regress both the people density and the scene depth. The pixels that have been attacked to alter the density estimate will also produce incorrect depths and can thus be detected.}
    \label{fig:deepsig}
\end{figure}

Fig.~\ref{fig:deepsig} depicts our approach, which we will refer to as the Density and Depth (DaD) model. We will show that it is robust to both unexposed attacks, in which the adversary does not know the existence of our adversarial detector, and exposed attacks, in which the adversary can access not only the density regression network but also our adversarial detector. In other words, even when our approach is exposed, a hacker cannot mount an effective attack while avoiding detection. This is largely because, 
depth measurements in a crowded scene, are affected by the appearance and disappearance of people but the perturbations remain small whereas those of RGB pixels can be much larger due to illumination and appearance changes. Therefore, even if the attacker has access to the depth maps, it remains difficult to guarantee that they will be altered in a consistent and undetectable way. 

Our contribution therefore is an effective approach to foiling adversarial attacks on people counting algorithms that rely on deep learning. Its principle is very generic and could be naturally extended to other image regression tasks. Our experiments on several benchmark datasets demonstrate that it outperforms heuristic-based and uncertainty-based attack detection strategies.


\section{Related Work}
\label{sec:related}

\vspace{0.2cm}
\parag{Crowd Counting.}

Early crowd counting methods~\cite{Wu05,Wang11a,Lin10} tended to rely on {\it counting-by-detection}, that is, explicitly detecting individual heads or bodies  and then counting them. Unfortunately, in very crowded scenes, occlusions make detection  difficult and these approaches have been largely displaced by {\it counting-by-density-estimation} ones~\cite{Zhang15c,Xiong17,Liu18c,Sindagi17,Xu19a,Onoro16,Shen18,Zhang16s,Sam17,Cao18,Wang19a,Liu19c,Liu19e,Shi19a,Liu19a,Liu19b}. They rely on training a regressor to estimate people density in various parts of the image and then integrating.  This trend began in~\cite{Chan09,Lempitsky10,Fiaschi12}, using either Gaussian Process or Random Forests regressors. Even though approaches relying on low-level features~\cite{Chen12f,Chan08,Brostow06,Rabaud06,Chan09,Idrees13} can yield good results, they have now mostly been superseded by CNN-based methods~\cite{Jiang19a,Zhao19a,Zhang19a,Wan19a,Lian19a,Liu19d,Yan19a,Ma19a,Liu19f,Xiong19a,Xu19a,Shi19b,Sindagi19a,Cheng19a,Wan19b,Zhang19b,Zhang19c},  a survey of which can be found in~\cite{Sindagi17}. In this paper, we therefore focus on attacks against these.

\parag{Defense against Adversarial Attacks.}

Deep networks trained to solve classification problems are vulnerable to adversarial attacks~\cite{Goodfellow2014a,Kurakin16,Madry18,Moosavi16,Moosavi17,Carlini17}. Existing attack strategies can be roughly categorized as optimization-based or gradient-based. The former~\cite{Moosavi16,Moosavi17,Carlini17}  involve terms related to the class probabilities, which makes the latter~\cite{Goodfellow2014a,Kurakin16,Madry18} better candidates for attacks against deep regression models. The very recent work of~\cite{Ranjan19a} is the only one we know of that examines adversarial attacks against a regressor, specifically one that estimates optical flow. However, it does not propose defense mechanisms, which are the focus of this paper.

In the context of classification, one popular defense is adversarial training~\cite{Szegedy13}, which augments the training data with adversarial examples and has been shown to outperform many competing methods~\cite{Grosse17,Gong17a,Metzen17,Bhagoji17a,Li17b,Feinman17a}. However, it needs access to adversarial examples, which are often not available ahead of time and must be generated during training. As a consequence, several alternative approaches have been proposed. This includes training auxiliary classifiers, ranging from simple linear regressors to complex neural networks, to predict whether a sample is adversarial or not. However, as shown in~\cite{Carlini17b}, such detection mechanisms can easily be defeated by an adversary targeting them directly. 

In any event, none of these methods are designed to detect attacks at the pixel-level. Even the few researchers who have studied adversarial attacks for semantic segmentation~\cite{Xiao18,Lis19}, which is a pixel-level prediction task, do not go beyond detection at the global image level. 

A seemingly natural approach to performing pixel-level attack detection would be to rely on prediction uncertainty. In~\cite{Feinman17a}, the authors argue that Bayesian uncertainty~\cite{Gal16} is a reliable way to detect the adversarial examples because the perturbed pixels generally have much higher uncertainty values. Uncertainty can be computed using dropout, as in~\cite{Gal16}, learned from data~\cite{Kendall17}, or estimated using the negative log-likelihood of each prediction~\cite{Lakshminarayanan17}. In our experiments, we will extend this strategy to pixel-wise adversarial attack detection and show that our approach significantly outperforms it.


\section{Density and Depth Model}
\label{sec:model}

As discussed in Section~\ref{sec:related}, most state-of-the-art crowd counting algorithms rely on a deep network regressor $\fd$, that takes an image $\bI$ as input and returns $D_{\bI}^{est} = \fd(\bI,\Theta)$, an estimated density map, which should be as close as possible to a ground-truth one $D_{\bI}^{gt}$ in $L^{2}$ norm terms. Here, $\Theta$ stands for the network's weights, that have been optimized for this purpose. 

An adversarial attack then involves generating a perturbation $\bP = \mathcal{F}_{p}(\bI,D_{\bI}^{c})$, where $\mathcal{F}_{p}$ maps the input image $\bI$ and the density $D_{\bI}^{c}$ associated with the clean image to $\bP$ in such a way that $\bA = \bI + \bP$ is visually indistinguishable from $\bI$ while yielding a crowd density estimate  $\fd(\bA,\Theta)$ that is as different as possible from the prediction obtained by the clean image $\bI$. 

We will review the best known ways to generate such attacks in Section~\ref{sec:attacks}. Here, our concern is to define $\fd$ so as to defeat them by ensuring that they are easily detected. To this end, we leverage an auxiliary task, depth estimation, as discussed below. 

\parag{Network Architecture.} 


\begin{figure*}[t]
\centering
 \includegraphics[width=1.0\linewidth]{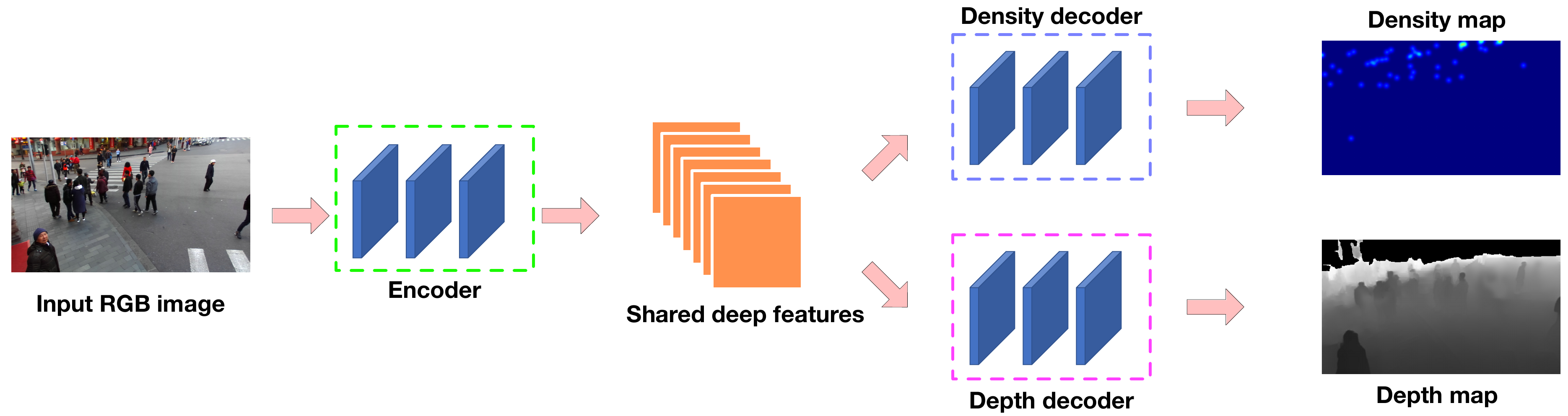}
\vspace{-3mm}
  \caption{ {\bf Density and Depth model.} An input RGB image is first encoded to deep features by an encoder network. Then, these features are decoded to a crowd density map and a depth map by two different decoder networks. At inference time, we detect adversarial attacks in a pixel-wise manner by observing the depth estimation errors.
}
  \label{fig:model}
  \end{figure*}

Instead of training a single regressor that predicts only people density, we train a two-stream network where one stream predicts people density and the other depth at each pixel. We write
\begin{align}
 D_{\bI}^{est} &= \fd(\bI,\Theta) \; , \label{eq:regress} \\
 Z_{\bI}^{est} &= \fz(\bI,\Theta)  \;.  \nonumber 
\end{align} 
where $D_{\bI}^{est}$ and $Z_{\bI}^{est}$ are the estimated densities and depths while $\fd$ and $\fz$ are two regressors parameterized by the weigths $\Theta$. The network that implements $\fd$ and $\fz$ comprises a single encoder and two decoders, one that produces the densities and the other the depths. It is depicted by Fig.~\ref{fig:model}, and we provide details of its architecture in the supplementary material. Note that some of the weights in $\Theta$ are specific to the first decoder and others to the second.

As the two decoders use the same set of features as input, it is difficult to tamper with the results of one without affecting that of the other, as we will see in Section~\ref{sec:compare}. More specifically, if a pixel is perturbed to change the local density estimate, the local depth estimate is likely to be affected as well. We therefore take the relative error in depth estimation with respect to a reference depth map $\bZ^{ref}$
\begin{equation}
 \frac{|\bZ^{est}(j)-\bZ^{ref}(j)|}{\bZ^{ref}(j)}
 \label{eq:indicator}
\end{equation}
to be an indicator of a potential disturbance. As will be shown in our experiments, the reference depth map can be either the ground-truth one, a rough estimate obtained using knowledge of the scene geometry, or the output of another monocular depth estimation network. In practice, we label a pixel as potentially tampered with if this difference is larger than  5\% of the largest difference in the training dataset and we will justify this choice in Section~\ref{sec:sensitivity}.  In test sequences for which the reference depth map can also be tampered with by the attacker, we can use the statistics of the training depth maps to also detect such tampering, as will be shown in Section~\ref{sec:robust}.

\parag{Network Training.} 

Given  a set of $N$ training images $\{\bI_{i}\}_{1 \leq i \leq N}$ with corresponding ground-truth density maps $\{\bD_{i}^{gt}\}_{1 \leq i \leq N}$ and ground-truth depth maps $\{\bZ_{i}^{gt}\}_{1 \leq i \leq N}$, we learn the weights $\Theta$ of the two regressors by minimizing the loss
\begin{small}
\begin{eqnarray}
    L(\Theta)  &=&  L_d+\lambda \cdot L_z \;, \\ \nonumber
    L_d &=& \frac{1}{2B}\sum_{i=1}^{B}\|\bD^{gt}_{i}-\bD^{est}_{\bI_i}\|^{2}_{2} \;, \\ \nonumber
    L_z &=& \frac{1}{2B}\sum_{i=1}^{B}\|\bZ^{gt}_{i}-\bZ^{est}_{\bI_i}\|^{2}_{2}\;.
    \label{eq:loss}
\end{eqnarray}
\end{small}
where $B$ is the batch size and $\lambda$ is a hyper-parameter that balances the contributions of the two losses. We found empirically that $\lambda=0.01$ yields the best overall performance, as will be shown in Section~\ref{sec:compare}. 

To obtain the ground-truth density maps $D^{gt}_{i}$, we rely on the same strategy as previous work~\cite{Li18f,Sam17,Zhang16s,Sam18,Liu18a}. In each training image $\bI_{i}$, we annotate a set of $c_{i}$ 2D points $O_{i}^{gt} = {\{O_{i}^j} \}_{1 \leq j \leq c_i}$ that denote the position of each human head in the scene. The corresponding ground-truth density map $\bD_{i}^{gt}$ is obtained by convolving an image containing ones at these locations and zeroes elsewhere with a Gaussian kernel  of mean $\mu$ and variance $\sigma$.


\newcommand{\ours}[0]{{\bf OURS}}
\newcommand{\rndh}[0]{{\bf RANDHALF}}
\newcommand{\rndq}[0]{{\bf RANDQUARTER}}
\newcommand{\uncertain}[0]{{\bf HETERO}}
\newcommand{\dropout}[0]{{\bf BAYESIAN}}
\newcommand{\simple}[0]{{\bf ENSEMBLES}}

\section{Experiments}
\label{sec:results}

In this section, we first introduce the existing adversarial attack methods that can be used against a deep regressor and describe the evaluation metric and the benchmark datasets we used to assess their performance. We then use them against our approach to demonstrate its robustness, and conclude with an ablation study that demonstrates that our approach is robust to the hyper-parameter setting and works well when used in conjunction with several recent crowd density regressors~\cite{Zhang16s,Li18f,Liu19a} .

\subsection{Attacking a Deep Regressor}
\label{sec:attacks}

While there are many adversarial attackers~\cite{Goodfellow2014a,Kurakin16,Madry18,Moosavi16,Moosavi17,Carlini17}, their effectiveness has been proven mostly against classifiers but far more rarely against regressors~\cite{Ranjan19a}. As discussed in Section~\ref{sec:related}, the gradient-based methods~\cite{Goodfellow2014a,Kurakin16,Madry18} are the most suitable ones to attack regressors, and we focus on the so-called  Fast Gradient Sign Methods ({\bf FGSM}s), which are the most successful and widely used ones. We will distinguish between unexposed attacks in which the attacker does not know that we use depth for verification purposes and exposed attacks in which they do.

\parag{Unexposed Attacks.} If the attacker is unaware that we use the depth map for verification purposes, they will only try to affect the density map.  They might then use one of the following variants of {\bf FGSM}. 

\begin{figure*}[t]
\centering
\begin{tabular}{ccccc}
 \includegraphics[width=.19\linewidth]{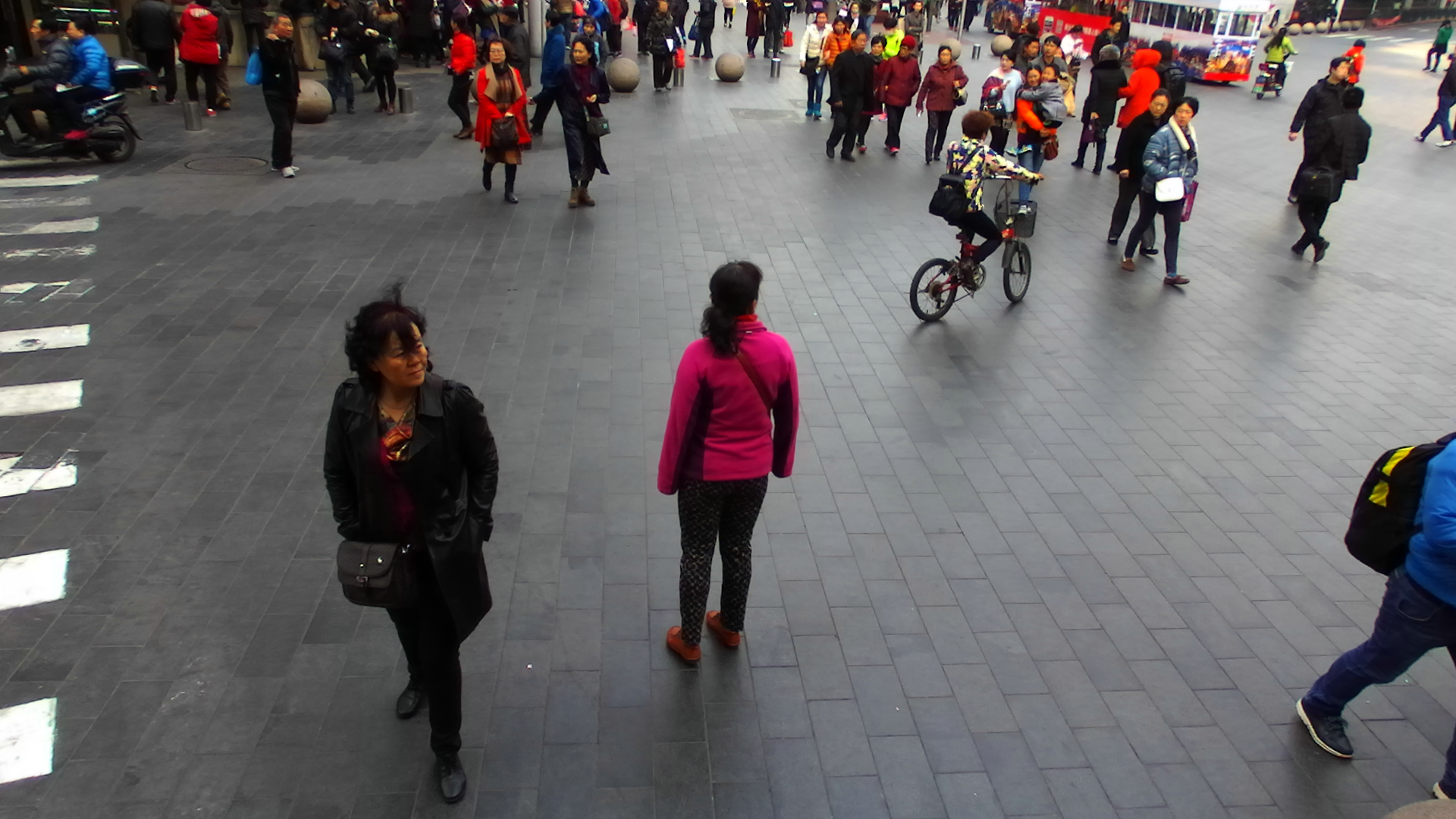}&
 \includegraphics[width=.19\linewidth]{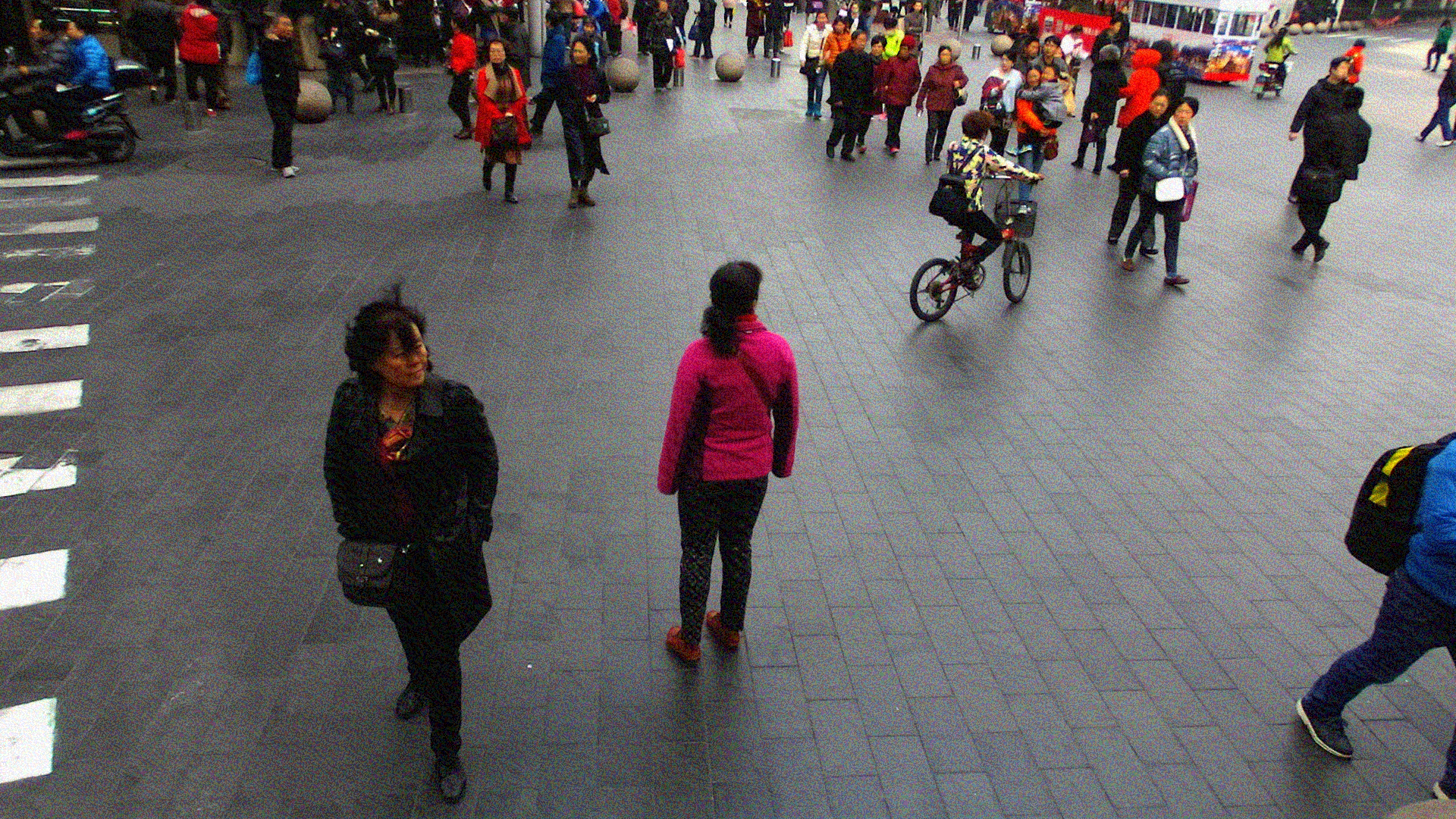}&
 \includegraphics[width=.19\linewidth]{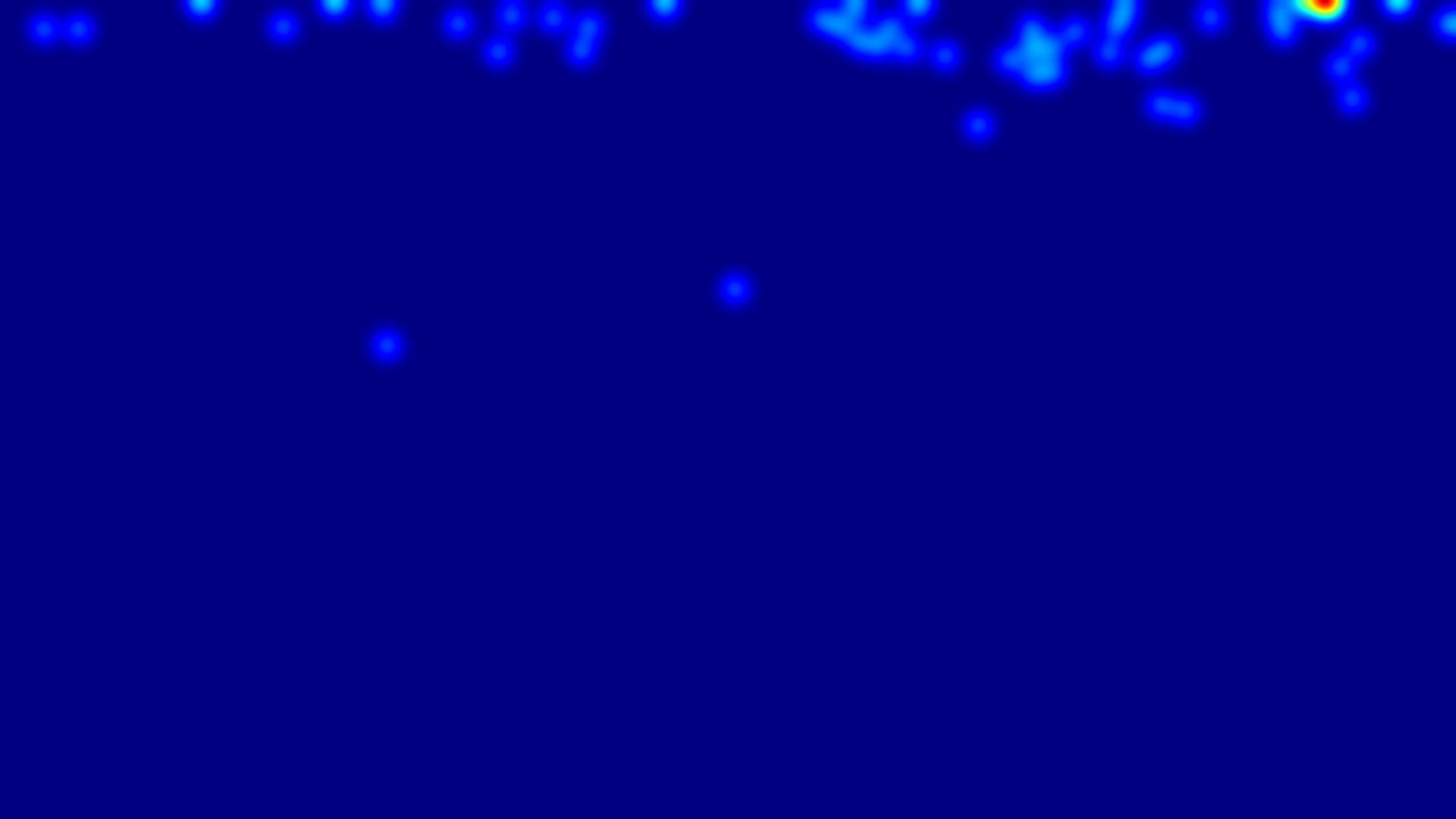}&
 \includegraphics[width=.19\linewidth]{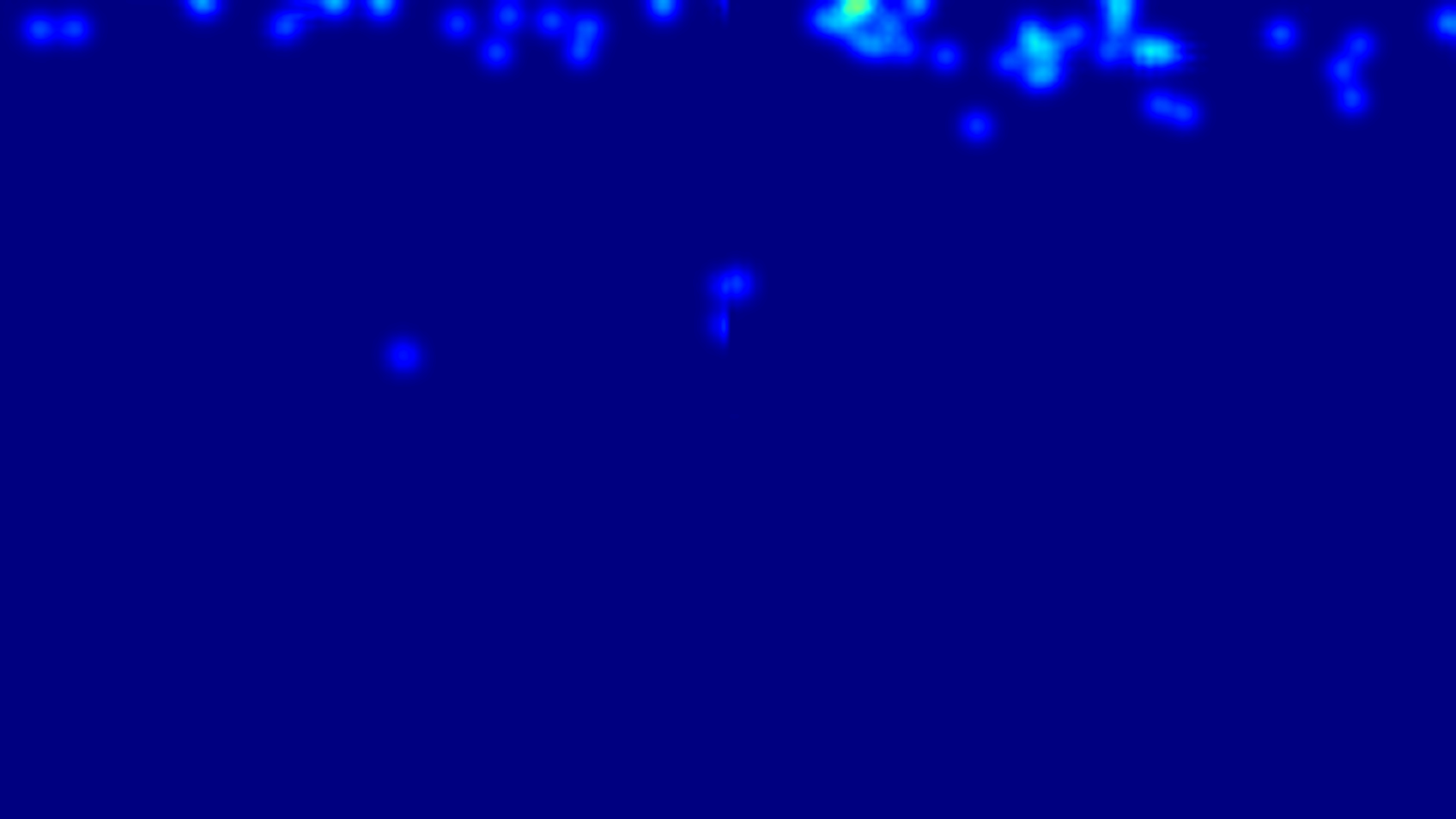}&
 \includegraphics[width=.19\linewidth]{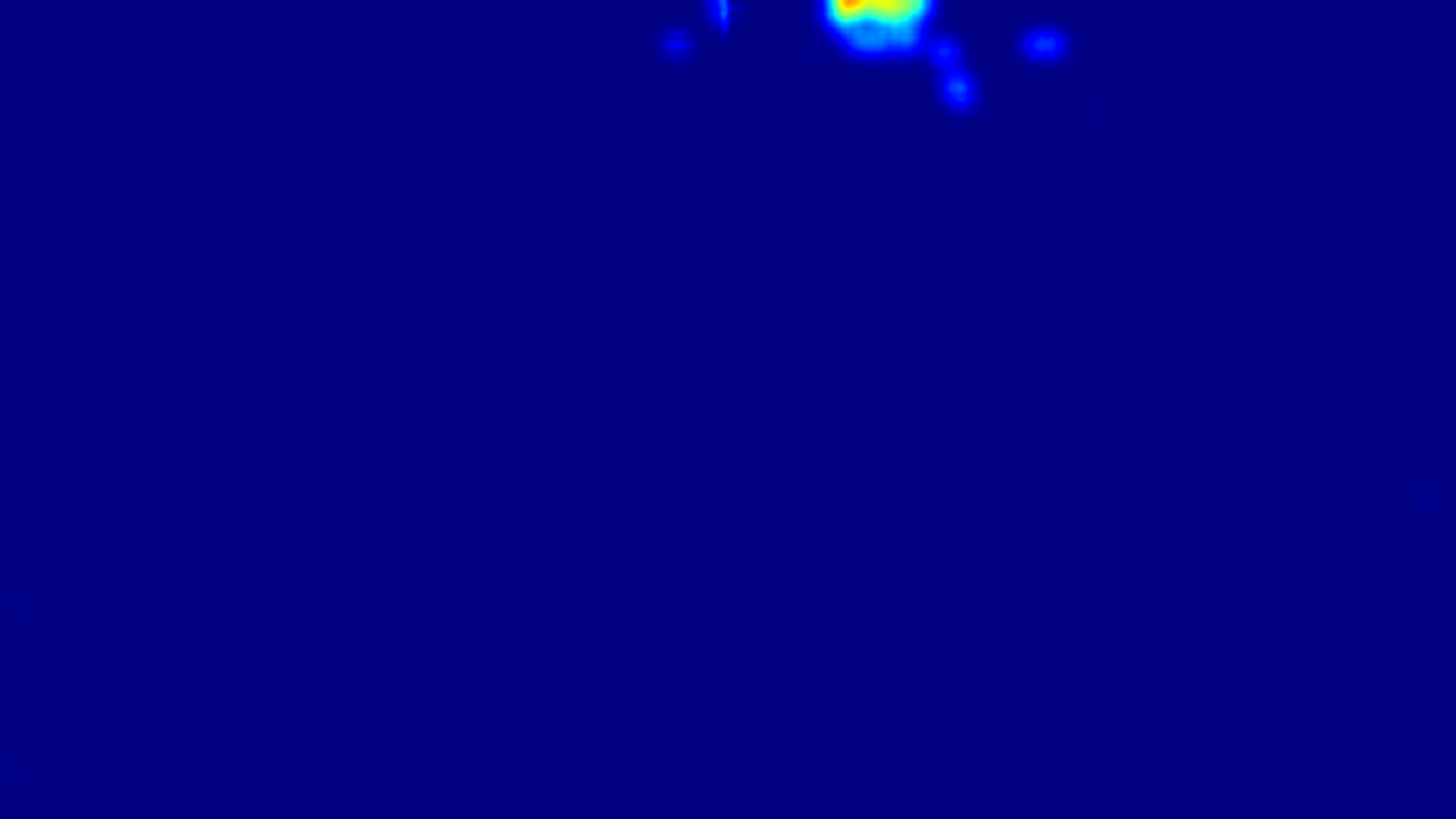}\\[-1mm]
 \hspace{-4mm}\footnotesize{(a)}&
 \hspace{-4mm}\footnotesize{(b)} &
 \hspace{-4mm}\footnotesize{(c)}&
 \hspace{-4mm}\footnotesize{(d)}& 
 \hspace{-4mm}\footnotesize{(e)}
\end{tabular}
\vspace{-3mm}
  \caption{ {\bf Crowd density estimation with original and perturbed images.} {\bf (a)} Original image. {\bf (b)} Image under \FGSMU{19} attack. {\bf (c)} Ground-truth density map with 51 people. {\bf (d)} Density map inferred from the original image, leading to an estimate of 51.8 people. {\bf (e)} Density map inferred from the perturbed image {\bf (b)}, yielding an estimated number of people of 18.1. Note the mismatch in density map and people counts between the original image and the perturbed one.
 }
  \label{fig:density}
  \end{figure*}

\vspace{0.1cm}
\sparag{\noindent (1) Untargeted  FGSM (\FGSMU{n})~\cite{Goodfellow2014a,Kurakin16}.}  It generates adversarial examples designed to increase the network loss as much as possible from that of the prediction obtained from the clean image, thereby preventing the network from predicting it. Given an input image $\bI$, the density predicted from the clean image $\bD$, and the regressor $\fd$ of Eq.~\ref{eq:regress} parametrized by $\Theta$, the attack is performed by iterating
\begin{small}
\begin{align}
    \bI_{0}^{adv} &  =  \bI \;   , \label{eq:FGSMu} \\
    \bI_{i+1}^{adv} &  = clip(\bI_{i}^{adv} + \alpha \cdot sign(\nabla_{\bI}L_d(\fd(\bI_{i}^{adv};\Theta),\bD)),\epsilon) \;   . \nonumber
\end{align}
\end{small}
$n$ times. The adversarial example is then taken to be $\bI_{n}^{adv}$, and we will refer to this as \FGSMU{n}. It is a single-step or multiple-step attack without target and $clip$ guarantees that the resulting perturbation is bounded by $\epsilon$. 
For consistency with earlier work~\cite{Goodfellow2014a,Kurakin16}, when  $n=1$, we reformulate this attack as
\begin{small}
    \begin{align}
        \bI^{adv} &  = \bI + \epsilon \cdot sign(\nabla_{\bI}L_d(\fd(\bI;\Theta),\bD)).   \label{eq:FGSMu1} 
    \end{align}
    \end{small}
Unless otherwise specified we use  $\epsilon=15$, $\alpha=1$, and $n=19$, as recommended in earlier work~\cite{Kurakin16}. These numbers are chosen to substantially increase the crowd counting error while keeping the perturbation almost imperceptible to the human eye. We will analyze the sensitivity of our approach to these values in Section~\ref{sec:sensitivity}. An example of this attack is shown in Fig.~\ref{fig:density}. By comparing Fig.~\ref{fig:density}(d) and (e), we can see that the attack made some people ``disappear''.


\vspace{0.1cm}
\sparag{\noindent (2) Targeted  FGSM (\FGSMT{n})~\cite{Kurakin16}.} Instead of simply preventing the network from finding the right answer $\bD$, we can target a specific wrong answer $\bD_{t}$. This is achieved using the slightly modified iterative scheme
\begin{small}
    \begin{align}
    \bI_{0}^{adv}    &  =  \bI  \; , \label{eq:FGSMt}\\ 
    \bI_{i+1}^{adv} &  =  clip(\bI_{i}^{adv} - \alpha \cdot sign(\nabla_{\bI}L_d(\fd(\bI_{i}^{adv};\Theta),\bD_{t})),\epsilon). \nonumber
\end{align}
\end{small}
Again, we take the adversarial example to be $\bI_{n}^{adv}$ and use the same values as before for $\epsilon$ and $\alpha$. We will refer to this as \FGSMT{n}. In our experiments, we take the targets to be the original density values plus one, which creates an obvious error while yielding tampered images that are undistinguishable from the original ones.

\parag{Exposed Attacks.} If the attacker knows that we are using the depth maps and has access to both $\fd$ and $\fz$, the two regressors of Eq.~\ref{eq:regress}, their natural reaction will be to try to modify the density maps while leaving the depth maps as unchanged as possible. To this end, we propose the following {\it exposed} variations of the  untargeted and targeted {\bf FGSM} attacks described above.

\sparag{\noindent (1) Untargeted Exposed FGSM (\FGSMUE{n}).} The iterative scheme becomes 
\begin{small}
    \begin{align}
    \bI_{0}^{adv} & = \bI \; , \label{eq:FGSMue}\\ 
    L_{i}^{all} & =     L_d(\fd(\bI_{i}^{adv};\Theta),\bD)- \lambda \cdot L_z(\fz(\bI_{i}^{adv};\Theta),\z)  \; , \nonumber \\ 
    \bI_{i+1}^{adv} & =     clip(\bI_{i}^{adv} + \alpha \cdot sign(\nabla_{\bI}L_{i}^{all}),\epsilon) \; . \nonumber
\end{align}
\end{small}
where $\z$ is the depth map associated with clean image. When $n=1$, we reformulate the final line of Eq.~\ref{eq:FGSMue} as in Eq.~\ref{eq:FGSMu1} for consistency with earlier work. The additional term in the loss function aims to preserve the predicting power of $\fz$ while compromising that of $\fd$ as much as possible. We again use the same values as before for $\epsilon$ and $\alpha$, and $\lambda=0.01$ is the same balancing factor as in Eq.~\ref{eq:loss}.

\sparag{\noindent (2) Targeted Exposed FGSM (\FGSMTE{n}).} Similarly, the targeted attack iterative scheme becomes
\begin{small}
    \begin{align}
    \bI_{0}^{adv} & =    \bI \;  \label{eq:FGSMut}\\ 
    L_{i}^{all} & =     L_d(\fd(\bI_{i}^{adv};\Theta),\bD_{t})+ \lambda \cdot L_z(\fz(\bI_{i}^{adv};\Theta),\z)   \; ,   \nonumber \\
    \bI_{i+1}^{adv} & =     clip(\bI_{i}^{adv} - \alpha \cdot sign(\nabla_{\bI}L_{i}^{all}),\epsilon) \; .  \nonumber
\end{align}
\end{small}
When $n=1$, we again reformulate the final line of Eq.~\ref{eq:FGSMut} as in Eq.~\ref{eq:FGSMu1}.

\subsection{Evaluation Datasets}

We use three different datasets to evaluate our approach. The first two are RGB-D datasets with ground-truth depth values obtained from sensors. Since depth sensors may not always be available, we also evaluate our model on a third dataset that contains RGB images with accurate {\it perspective maps}, that is, maps inferred from the scene geometry without using depth sensors. Such maps only represent the scene to the exclusion of the people in it. This will let us show that
our approach not only works for RGB-D datasets but also achieves remarkable performance in RGB images if scene geometry is available. Furthermore, we will show that our approach also applies when the reference depth is obtained using a separate monocular depth estimation network.

\parag{ShanghaiTechRGBD~\cite{Lian19a}.} This is a large-scale RGB-D dataset with 2,193 images and 144,512 annotated heads. The valid depth ranges from 0 to 20 meters due to the limitation in depth sensors. The lighting condition ranges from very bright to very dark in different scenarios. We use the same setting as in~\cite{Lian19a}, with 1,193 images as training set and the remaining ones as test set,  and normalize the depth values from [0,20] to [0,1] for both training and evaluation.

\parag{MICC~\cite{Bondi14a}.} This dataset was acquired by a fixed indoor surveillance camera. It is divided into three video sequences, named FLOW, QUEUE and GROUPS. The crowd motion varies from sequence to sequence. In the FLOW sequence, people walk from point to point. In the QUEUE sequence, 
people walk in a line. In the GROUPS sequence, people move inside a controlled area. There are 1,260 frames in the FLOW sequence with 3,542 heads. The QUEUE sequence contains 5,031 heads in 918 frames, and the GROUPS sequence encompasses 1,180 frames with 9,057 heads.
We follow the same setting as in~\cite{Lian19a}, taking 20\% of the images of each scene as training set and using the remaining ones as test set.

\parag{Venice~\cite{Liu19a}.} The above two RGB-D datasets contain depth information acquired by sensors. Such information is hard to obtain in outdoor environments, particularly if the scene is far from the camera. Therefore, we also evaluate our approach on the \textit{Venice} dataset. This dataset contains RGB images and an accurate perspective map of each scene. It was inferred from the grid-like ground pattern that  is provided in the supplementary material and without using a depth sensor. The dataset contains 4 different sequences for a total of 167 annotated frames with fixed 1,280 $\times$ 720 resolution. Our experimental setting follows that of~\cite{Liu19a,Liu19b}, with 80 images from a single long sequence as training data, and the images from the remaining 3 sequences for testing purposes.

\subsection{Metrics and Baselines}

In all our experiments, we partition the images into four parts and tamper with one while leaving the other three untouched. We then measure:
\begin{itemize}

  \item How well we can detect the pixels that have been tampered with.  We measure this in terms of the mean Intersection over Union
\begin{equation}
    mIoU = \frac{1}{N} \sum_{i=1}^{N} \frac{ |v_{i} \cap \hat{v}_{i}|}{ | v_{i} \cup \hat{v}_{i} |}
\end{equation}
where $N$ is the number of images, $\hat{v}_{i}$ is 1 for the pixels in image $\bI_i$ predicted to have been tampered with according to Eq.~\ref{eq:indicator} and $v_{i}$ is the ground-truth perturbation mask.
  
  \item How well the modifications of the depth map correlate with those of the predicted density.  As in many previous works~\cite{Zhang16s,Zhang15c,Onoro16,Sam17,Xiong17,Sindagi17,Liu19a,Liu19b}, we quantify these modifications in terms of the mean absolute error for densities and depths along with the root mean squared error for density. They are defined as  
\begin{align}
    DMAE & = \frac{1}{N}\sum_{i=1}^{N}|d_{i}-\hat{d_{i}}|\; , \nonumber \\
    ZMAE  & = \frac{1}{N}\sum_{i=1}^{N}\frac{ \sum_{j=1}^{M_{i}} |z_{i}^{j}-\hat{z_{i}^{j}}|}{M_{i}}  \;, \\
    RMSE & = \sqrt{\frac{1}{N}\sum_{i=1}^{N}(d_{i}-\hat{d_{i}})^{2}}\; . \nonumber  
\end{align}
where $N$ is the number of test images, $d_{i}$ and $z_{i}^{j}$ denote the true number of people in the $i$th image and depth value at pixel $j$ of the $i$th image, and $\hat{d_{i}}$ and $\hat{z_{i}^{j}}$ are the estimated values. $M_{i}$ is the number of tampered pixels in the $i$th image. In practice $\hat{d_{i}}$ is obtained by integrating the predicted people densities. 

\end{itemize}
In the absence of prior art on defenses against attacks of density estimation algorithms, we use the following baselines for comparison purposes. 
\begin{itemize}

\item \rndh{} and \rndq{}. We randomly label either half or a quarter of the pixels as being under attack, given that we know {\it a priori} that exactly a quarter are.  We introduced  \rndh{} to show that using a random rate other than the true one does not help. 
 
\item \uncertain{}. Since an adversarial attack is caused by modifying the input image, it can be seen as heteroscedastic aleatoric uncertainty~\cite{Kendall17}, which assumes that observation noise varies with the input, that is, uncertainty caused only by the input and invariant to the model. We threshold the uncertainty values to classify each pixel as perturbed or not and report the results obtained with the best threshold. 

\item \simple{}. We use the approach of~\cite{Lakshminarayanan17} that relies on a scalable method for estimating predictive uncertainty from deep networks using a scoring rule as training criterion. The optional adversarial training process is not used as we do not know the potential attackers in advance. As before, we threshold the uncertainty values to obtain a pixel-wise classification map and report the best results.

\item \dropout{}. We further compare our model with Bayesian uncertainty~\cite{Gal16}, which uses dropout to
approximate model uncertainty. Again, we threshold the uncertainty value and report the results for the best threshold. 
\end{itemize}

The baseline models are trained with the same backbone as our approach.

\subsection{Comparative Performance}
\label{sec:compare}

\paragraph{Using the CAN~\cite{Liu19a} architecture.} 

CAN is an encoder-decoder crowd density estimation architecture that delivers excellent performance. We use it to implement $\fd$ and duplicate its decoder to implement $\fz$.  Recall from Section~\ref{sec:model} that the hyper-parameter $\lambda$ of Eq.~\ref{eq:loss} balances the people density estimation loss and the depth estimation one while training $\fd$ and $\fz$. In Table~\ref{tab:lambda}, we report the errors of the two regressors as a function of  $\lambda$. $\lambda=0.01$ yields the best performance overall, and we use regressors trained using this value in all our other experiments. Interestingly, training $\fd$ and $\fz$ jointly yields a better density regressor than training $\fd$ alone, which is what we do when we set $\lambda$ to zero. 


\begin{table}[t]
  \centering
  \scalebox{0.8}{
    \rowcolors{2}{white}{gray!10}
    \begin{tabular}{lccccccccc}
      \toprule
    &\multicolumn{3}{c}{ShanghaiTechRGBD}& \multicolumn{3}{c}{MICC}& \multicolumn{3}{c}{Venice} \\
  $\lambda$   & $DMAE$ & $RMSE$ & $ZMAE$& $DMAE$ & $RMSE$ & $ZMAE$& $DMAE$ & $RMSE$ & $ZMAE$ \\
  \midrule
  0.0  & 4.82 & 7.23 &  NA & 0.91 & 0.98 &  NA & 23.51 & 38.92 & NA \\
  0.001  & 4.76 & 7.19 & 0.21 & 0.86 & 0.93 & 2.26 & {\bf 21.81} & 24.91 & 2.59  \\
  0.01  & {\bf 4.32} & {\bf 7.16} & 0.04 & {\bf 0.52}& {\bf 0.67} & {\bf 1.36}& 21.92 & {\bf 24.74} & {\bf 1.13}  \\ 
  0.1  & 4.61 & 7.41 & {\bf 0.03} & 0.61& 0.73 & 1.43 & 23.12 & 26.52 & 1.23  \\
  1.0  & 4.80 & 7.26 & 0.04 & 0.89 & 0.93 & 1.47 & 23.27 &  32.16 & 1.32  \\
  10.0  & 4.92 & 8.01 & 1.16 & 0.98 & 1.04 & 1.72 & 25.43 & 39.65 & 1.86  \\
  \bottomrule
  \end{tabular}
  }
  \caption{{\bf Error summary of crowd density and depth estimation for different $\lambda$ values.}}
  \label{tab:lambda}
\end{table}


\begin{table}[t]
  \centering
  \scalebox{0.8}{
    \rowcolors{2}{white}{gray!10}
    \begin{tabular}{lccccccccc}
      \toprule
    &\multicolumn{3}{c}{ShanghaiTechRGBD}& \multicolumn{3}{c}{MICC}& \multicolumn{3}{c}{Venice} \\
  Attack  & $DMAE$ & $RMSE$ & $ZMAE$& $DMAE$ & $RMSE$ & $ZMAE$& $DMAE$ & $RMSE$ & $ZMAE$ \\
  \midrule
  Original image  & {\bf 4.32} & {\bf 7.16} & {\bf 0.04} & {\bf 0.52}& {\bf 0.67} & {\bf 1.36}& {\bf 21.92} & {\bf 24.74} & {\bf 1.13}  \\
  \FGSMU{1}  & 61.56 & 71.58 & 0.12 &2.45 & 3.01 & 7.46&  78.75 & 88.65 & 2.16 \\
  \FGSMT{1}& 60.31 & 70.08 & 0.12 & 1.66& 1.87 & \textcolor{red}{7.77}& \textcolor{red}{202.54} & \textcolor{red}{204.65} & \textcolor{red}{2.62} \\
  \FGSMU{19} & \textcolor{red}{64.55}& \textcolor{red}{75.11} & \textcolor{red}{0.14} & \textcolor{red}{3.13} & \textcolor{red}{3.75} & 7.68& 48.56 & 57.83 & 1.76 \\
  \FGSMT{19}& 62.86 & 73.09 & 0.13 & 1.90 & 2.15 & \textcolor{red}{7.77}& 112.17 & 115.24&  1.93\\
  \FGSMUE{1}  & 58.14  & 68.34 & 0.11 & 2.72 & 3.33 & 6.66 & 58.40& 66.94& 2.03\\ 
  \FGSMTE{1}& 53.64 & 63.43 & 0.11 & 2.47 & 3.02 & 6.65 & 171.76 &174.31 & 2.53 \\ 
  \FGSMUE{19}& 63.81 & 74.30 & 0.10 & 2.44& 2.97 & 5.26 & 42.74 & 51.20& 1.80 \\ 
  \FGSMTE{19}& 52.89 & 62.43 & 0.10 & 2.30& 2.81 & 5.26& 95.71 & 99.13& 1.92 \\ 
  \bottomrule
  \end{tabular}
  }
  \caption{{\bf Error summary of crowd density and depth estimation.}}
  \label{tab:error}
\end{table}

We report the counting and depth errors with/without attack in Table~\ref{tab:error} for the 3 datasets. All the attacks cause large increase in crowd counting errors, which always comes with a substantial increase in depth estimation error. The exposed methods reduce slightly this increase but at the cost of also making the attack less effective. 


\begin{table}
  \centering
  \scalebox{0.7}{
    \rowcolors{2}{white}{gray!10}
    \begin{tabular}{lcccccc}
      \toprule
  Attack & \rndh & \rndq &  \uncertain~\cite{Kendall17} & \simple~\cite{Lakshminarayanan17}&\dropout~\cite{Gal16}&   \ours \\
  \midrule
  \FGSMU{1} & 0.20& 0.14 &0.23 & 0.35& 0.23& {\bf 0.54}\\
  \FGSMT{1} & 0.20& 0.14 & 0.23 & 0.32 &0.24 & {\bf 0.54}\\
  \FGSMU{19}& 0.20& 0.14 &0.28 & 0.36& 0.23& {\bf 0.58}\\
  \FGSMT{19}& 0.20& 0.14 & 0.28 & 0.33 & 0.24& {\bf 0.57}\\
  \FGSMUE{1} & 0.20& 0.14 &0.24 & 0.28& 0.23& {\bf 0.52}\\
  \FGSMTE{1} & 0.20& 0.14& 0.21& 0.30 & 0.23& {\bf 0.51}\\
  \FGSMUE{19}& 0.20& 0.14 & 0.20& 0.33 & 0.23& {\bf 0.45}\\
  \FGSMTE{19}& 0.20& 0.14 & 0.25& 0.30& 0.24& {\bf 0.47}\\
  \bottomrule
  \end{tabular}}
  \caption{{\bf mIoU of pixel-wise adversarial detection on ShanghaiTechRGBD.}}
  \label{tab:miou_ShanghaiTechRGBD}
\end{table}

\begin{table}
  \centering
  \scalebox{0.7}{
    \rowcolors{2}{white}{gray!10}
    \begin{tabular}{lcccccc}
      \toprule
  Attack & \rndh & \rndq &  \uncertain~\cite{Kendall17} & \simple~\cite{Lakshminarayanan17}&\dropout~\cite{Gal16}&   \ours \\
  \midrule
  \FGSMU{1} & 0.20& 0.14 &0.30 & 0.35& 0.28& {\bf 0.46}\\
  \FGSMT{1} & 0.20& 0.14 & 0.33 & 0.32 &0.26 & {\bf 0.49}\\
  \FGSMU{19}& 0.20& 0.14 &0.30 & 0.30& 0.27& {\bf 0.49}\\
  \FGSMT{19}& 0.20& 0.14 & 0.32 & 0.37 & 0.23& {\bf 0.49}\\
  \FGSMUE{1} & 0.20& 0.14 &0.30 & 0.35& 0.26& {\bf 0.41}\\
  \FGSMTE{1} & 0.20& 0.14& 0.28& 0.31& 0.28& {\bf 0.41}\\
  \FGSMUE{19}& 0.20& 0.14 & 0.31& 0.33 & 0.28& {\bf 0.40}\\
  \FGSMTE{19}& 0.20& 0.14 & 0.30& 0.34& 0.27& {\bf 0.40}\\
  \bottomrule
  \end{tabular}}
  \caption{{\bf mIoU of pixel-wise adversarial detection on MICC.}}
  \label{tab:miou_micc}
\end{table}

\begin{table}
  \centering
  \scalebox{0.7}{
    \rowcolors{2}{white}{gray!10}
    \begin{tabular}{lcccccc}
      \toprule
   Attack & \rndh & \rndq &  \uncertain~\cite{Kendall17} & \simple~\cite{Lakshminarayanan17}&\dropout~\cite{Gal16}&   \ours \\
  \midrule
  \FGSMU{1} & 0.20& 0.14 &0.24 & 0.19& 0.23& {\bf 0.42}\\
  \FGSMT{1} & 0.20& 0.14 & 0.26 & 0.20 &0.22 & {\bf 0.49}\\
  \FGSMU{19}& 0.20& 0.14 &0.25 & 0.16& 0.22& {\bf 0.36}\\
  \FGSMT{19}& 0.20& 0.14 & 0.26 & 0.18 & 0.23& {\bf 0.38}\\
  \FGSMUE{1} & 0.20& 0.14 &0.22 & 0.20& 0.23& {\bf 0.40}\\
  \FGSMTE{1} & 0.20& 0.14& 0.23& 0.20& 0.25& {\bf 0.48}\\
  \FGSMUE{19}& 0.20& 0.14 & 0.26 & 0.21 & 0.24& {\bf 0.38}\\
  \FGSMTE{19}& 0.20& 0.14 & 0.22 & 0.19& 0.23& {\bf 0.41}\\
  \bottomrule
  \end{tabular}}
  \caption{{\bf mIoU of pixel-wise adversarial detection on Venice.}}
  \label{tab:miou_venice}
\end{table}

\begin{figure*}[t]
\centering
\begin{tabular}{ccccc}
 \includegraphics[width=.2\linewidth]{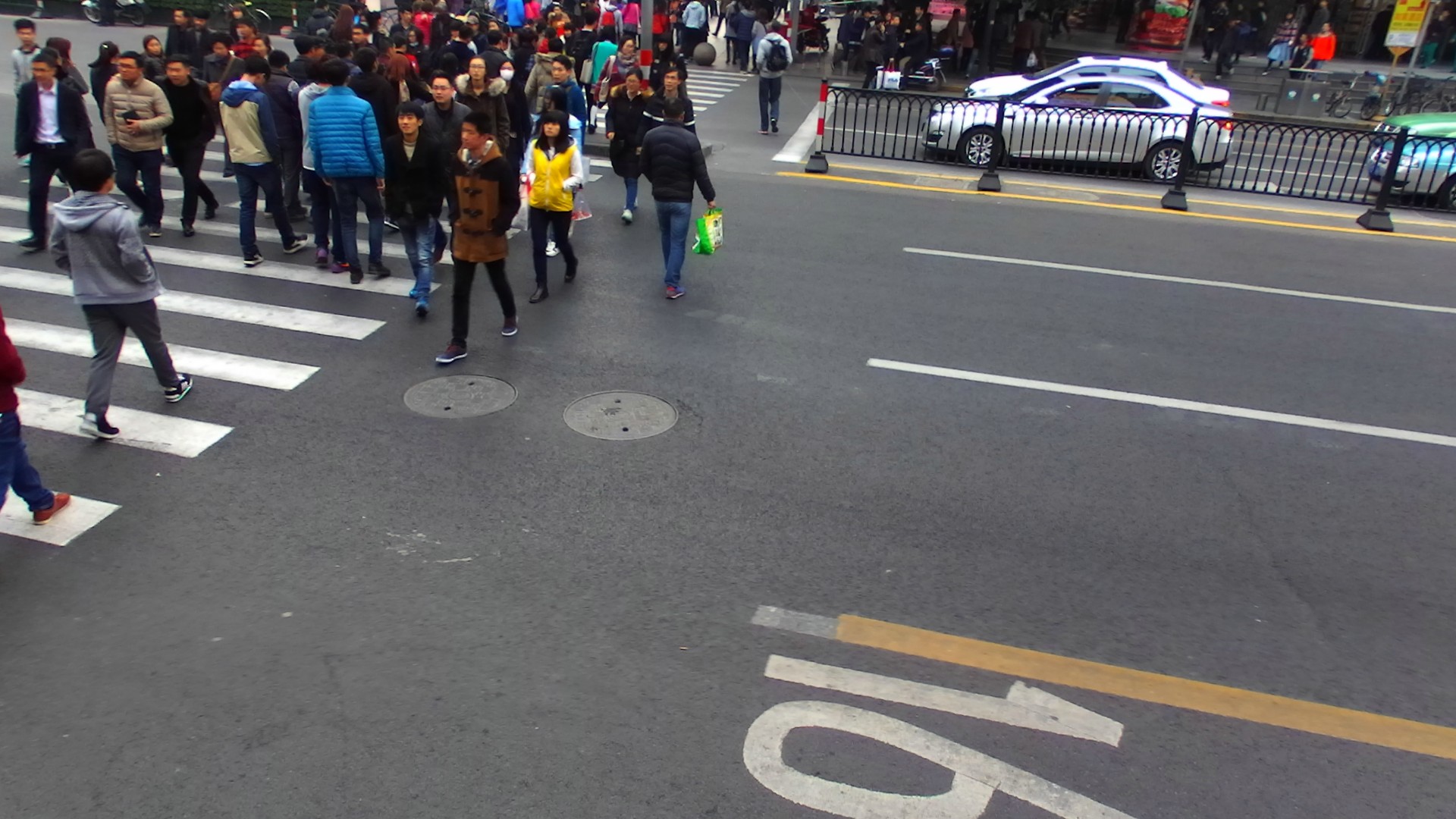}&
 \includegraphics[width=.2\linewidth]{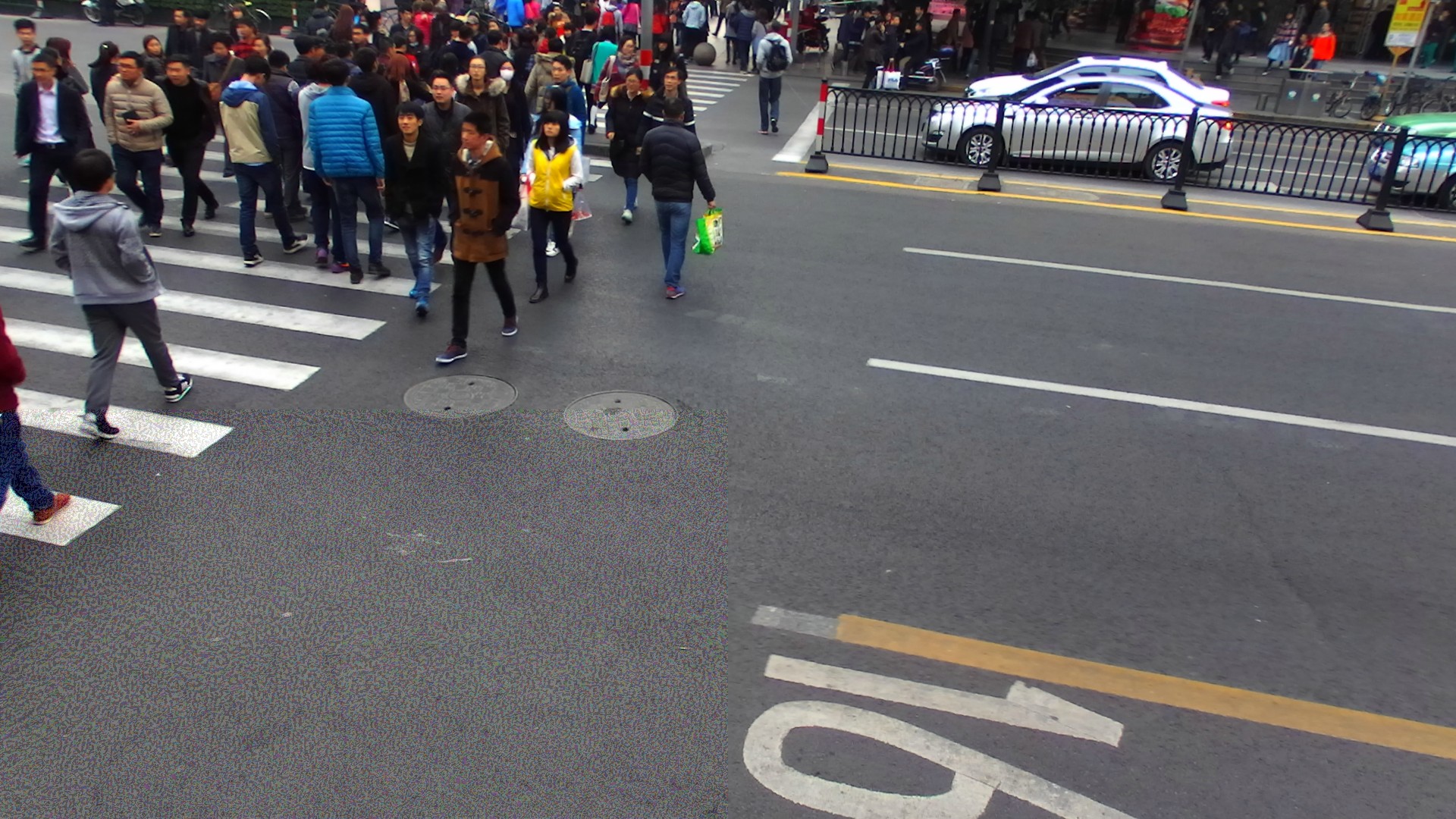}&
 \includegraphics[width=.2\linewidth]{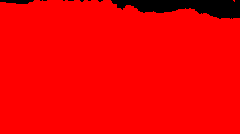}&
 \includegraphics[width=.2\linewidth]{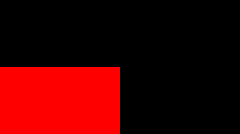}&
 \includegraphics[width=.2\linewidth]{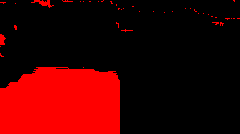}\\[-1mm]
\tiny{Original image}&
\tiny{Image under attack} &
\tiny{Region of interest}&
\tiny{Ground truth}& 
\tiny{\ours}
\end{tabular}
  \caption{ {\bf Pixel-wise adversarial detection on ShanghaiTechRGBD.} We show the original image, the image under \FGSMU{1} attack, the ROI(red), the ground-truth attacked region within the ROI(red), and the attacked region estimated by our method. Note how similar the attacked region mask produced by \ours{} is to the ground truth.
 }
  \label{fig:visualization}
  \end{figure*}

In Tables~\ref{tab:miou_ShanghaiTechRGBD}, \ref{tab:miou_micc}, and \ref{tab:miou_venice}, we report the pixel-wise adversarial detection accuracy for the \textbf{ShanghaiTechRGBD}, \textbf{MICC} and \textbf{Venice} datasets. Our approach outperforms all the baseline models by a large margin for all the attacks. In Fig.~\ref{fig:visualization}, we show a qualitative result. 


\begin{figure*}
\centering
\begin{tabular}{cccc}
 \includegraphics[width=.33\linewidth]{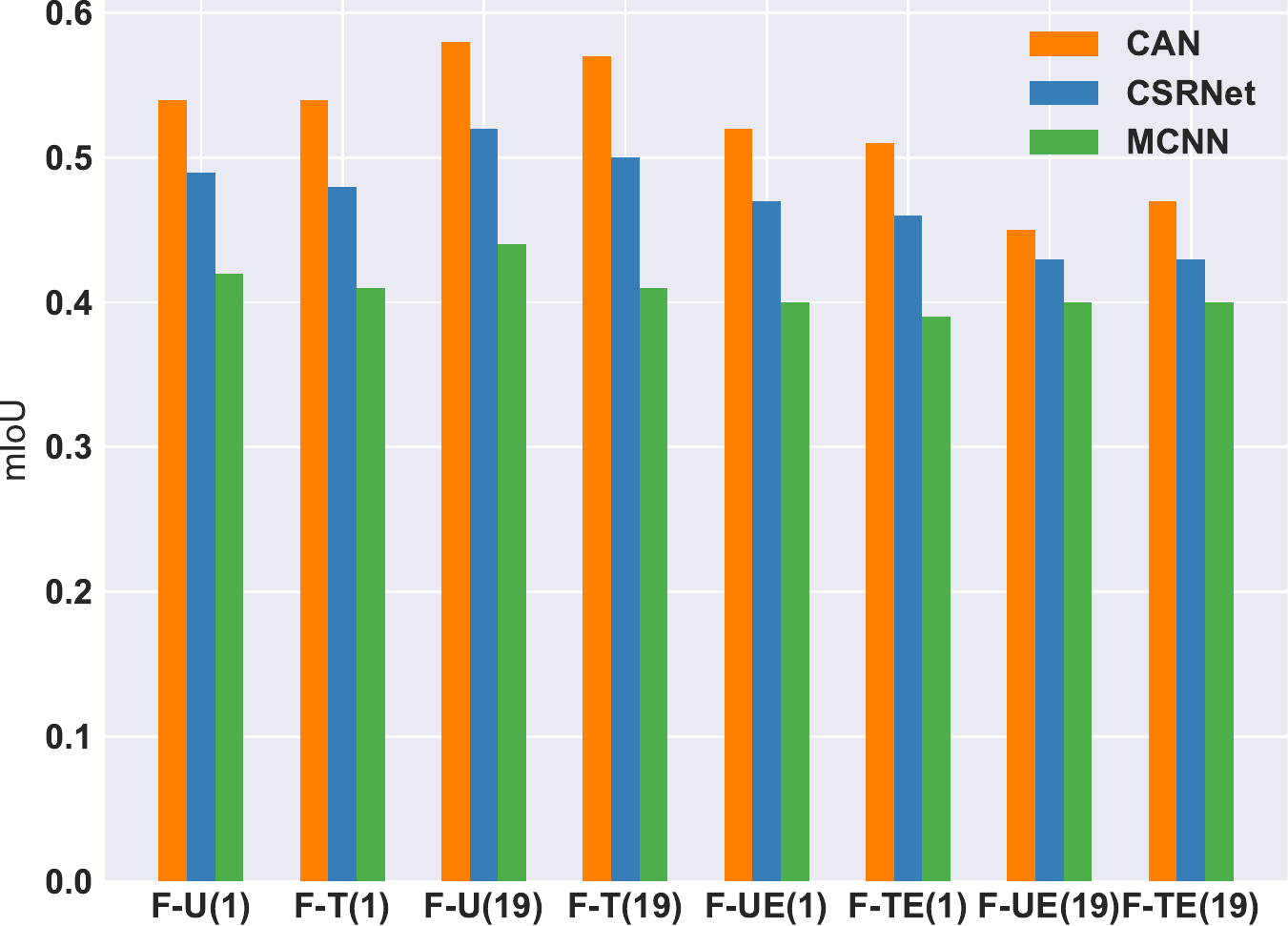}&
 \includegraphics[width=.33\linewidth]{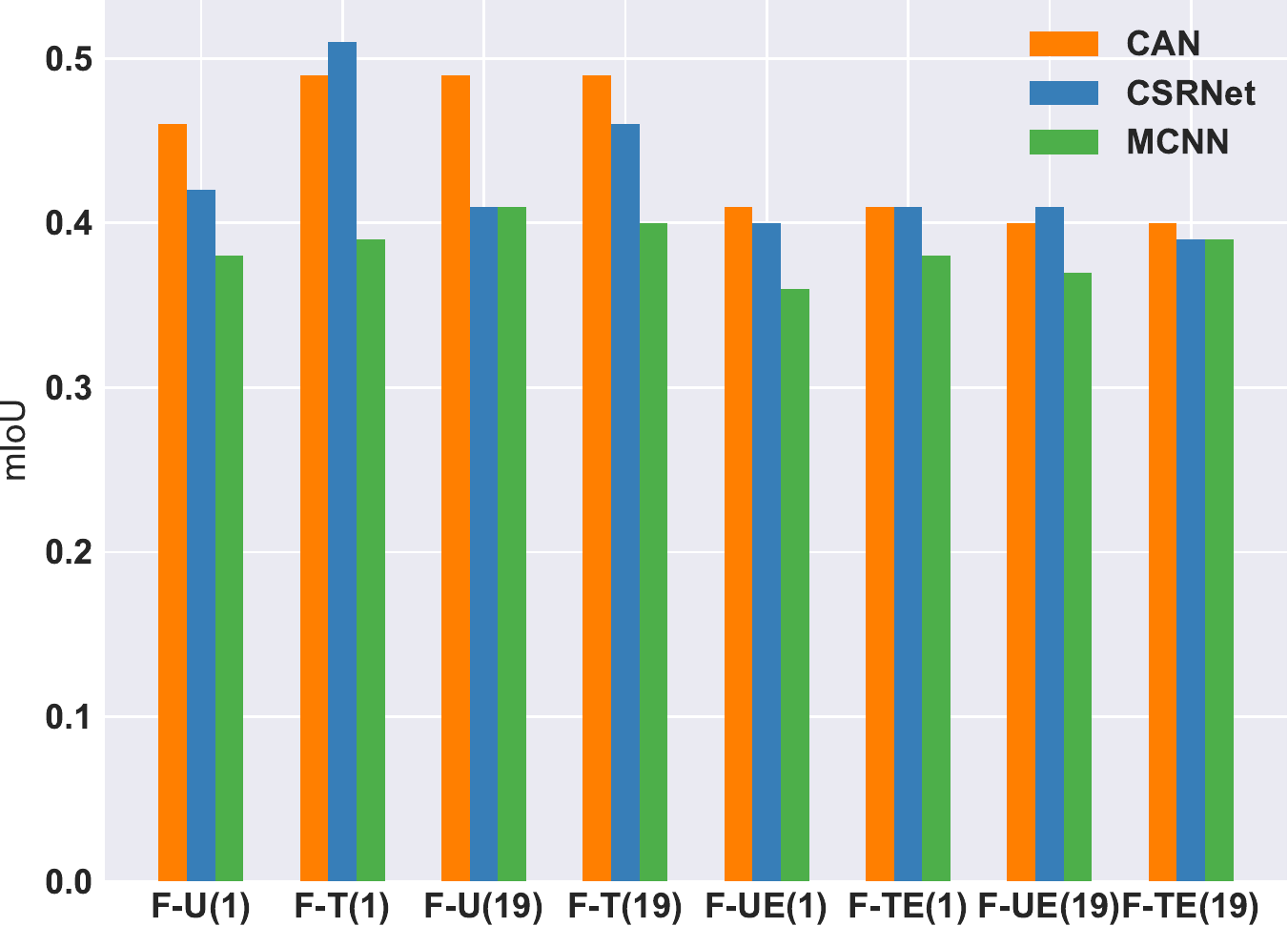}&
 \includegraphics[width=.33\linewidth]{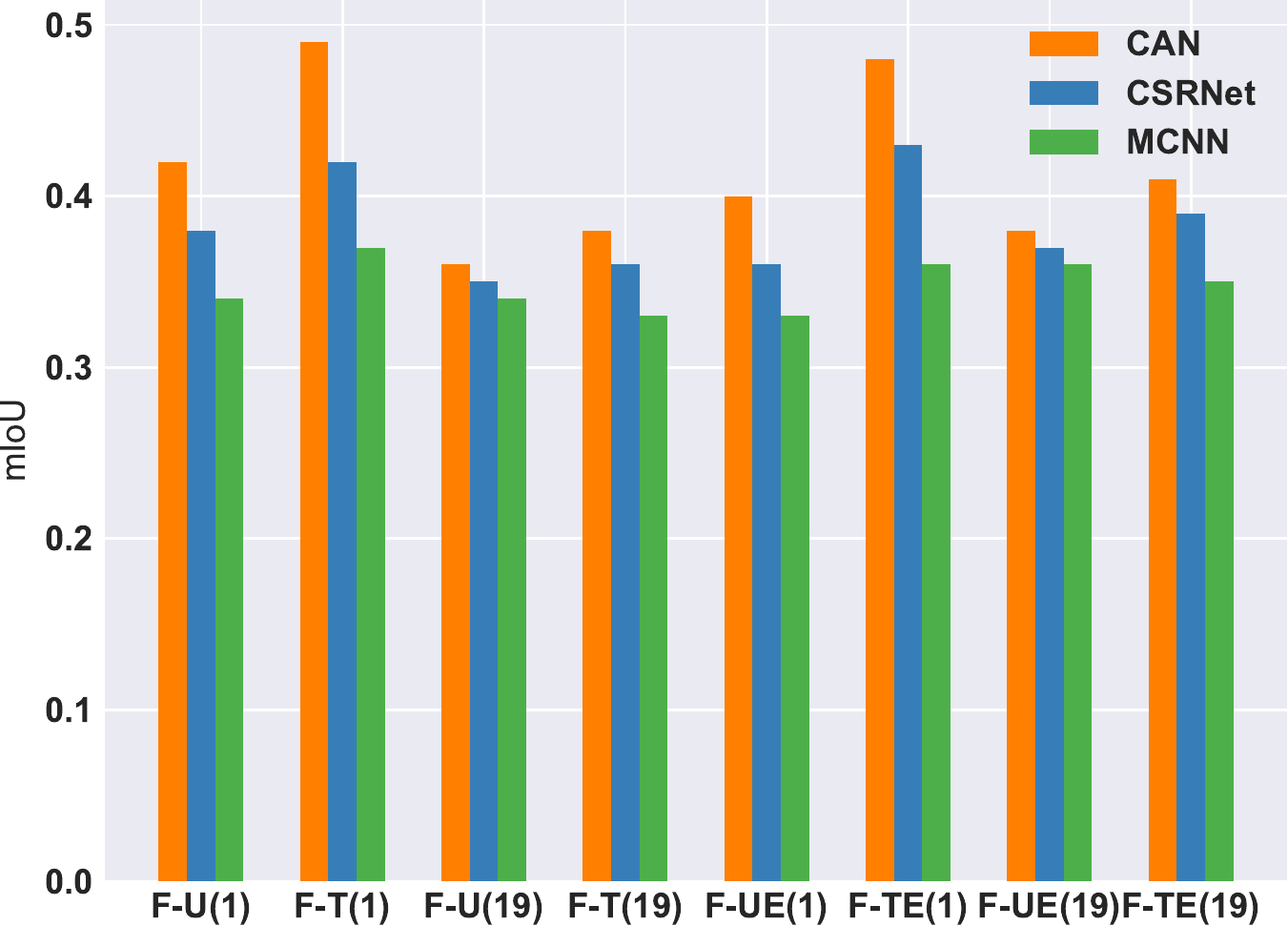}\\
 \hspace{-4mm}\footnotesize{{\bf ShanghaiTechRGBD } dataset}&
 \hspace{-4mm}\footnotesize{{\bf MICC } dataset} &
 \hspace{-4mm}\footnotesize{{\bf Venice } dataset}
\end{tabular}
\vspace{-3mm}
  \caption{ {\bf Detection accuracy with different backbones.} We report the mIoU of different backbones on different datasets.
}
  \label{fig:backbones}
  \end{figure*}


\begin{table}
  \begin{varwidth}[b]{0.4\linewidth}
  \centering
  \scalebox{0.7}{
    \rowcolors{2}{white}{gray!10}
    \begin{tabular}{lccc}
      \toprule
  Attack & ShanghaiTechRGBD & MICC & Venice  \\
  \midrule
  \FGSMU{1} & 0.48 & 0.42 & 0.40\\
  \FGSMT{1} & 0.48 & 0.44 & 0.44\\
  \FGSMU{19}& 0.50 & 0.46 & 0.33\\
  \FGSMT{19}& 0.51 & 0.45 & 0.37\\
  \FGSMUE{1} & 0.46 & 0.40 &0.40\\
  \FGSMTE{1} & 0.44 & 0.39 & 0.42\\
  \FGSMUE{19}& 0.42 & 0.38 &0.36\\
  \FGSMTE{19}& 0.43 & 0.39 &0.37\\
  \bottomrule
  \end{tabular}}
  \caption{{\bf mIoU of pixel-wise adversarial detection on different datasets using depth maps inferred by VNL~\cite{Yin19a}.}}
  \label{tab:miou_pretrain}
\end{varwidth}%
\hfill
\begin{minipage}[b]{0.5\linewidth}
  \centering
  \includegraphics[width=1.0\linewidth]{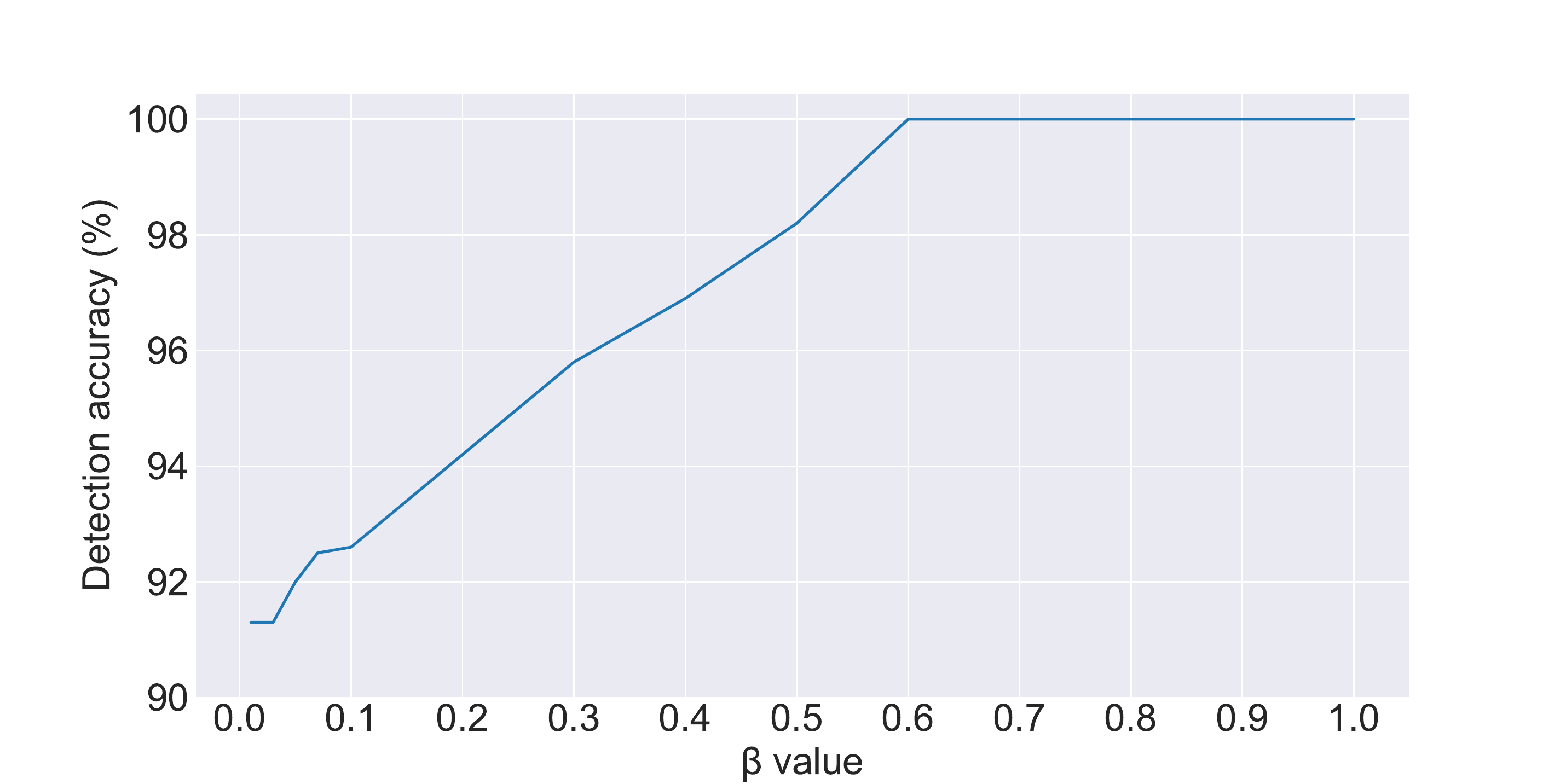}
  \vspace{-3mm}
    \captionof{figure}{ {\bf Detection accuracy on MICC.} We report the detection accuracy for depth values tampered with different strengths.
  }
    \label{fig:depth_detection}
  \end{minipage}
\end{table}


\begin{figure*}[htbp]
\centering
 \includegraphics[width=1.0\linewidth]{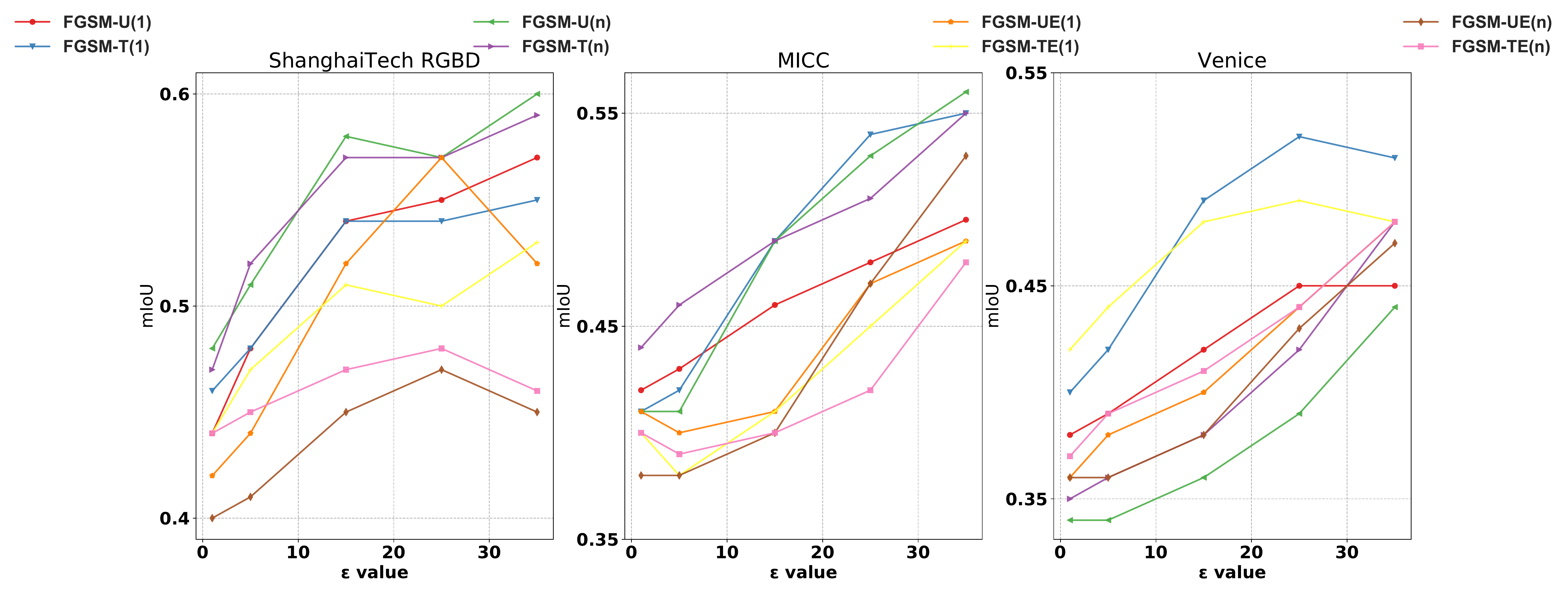}
\vspace{-3mm}
  \caption{ {\bf Detection accuracy with different perturbation strengths.} We report the mIoU for different $\epsilon$ values on different datasets.}
  \label{fig:strength}
  \end{figure*}

\parag{Using the  CSRNet~\cite{Li18f} and MCNN~\cite{Zhang16s} architectures.} 

To show that the above results are not tied to the CAN architecture, we re-ran our experiments using CSRNet~\cite{Li18f} and MCNN~\cite{Zhang16s}. As shown in Fig.~\ref{fig:backbones}, we obtain similar mIoU scores for all three architectures, with a slight advantage for the more recent CAN.

\subsection{Inference without Depth Map}
\label{sec:pretrain}

Using the scene geometry to infer a depth map, as we did in the \textit{Venice} dataset, is one way to avoid  having to use a depth-sensor to create a reference depth map. Unfortunately, the strong geometric patterns we used to do this are not present in all images. In this section, we therefore explore the use of VNL~\cite{Yin19a}, a pre-trained deep model that can estimate depth from single images, to create the reference depth map. We report our results in  
Table~\ref{tab:miou_pretrain}.  As could be expected, there is a slight performance drop compared to the results we obtained using the ground-truth depth map, given in  Tables~\ref{tab:miou_ShanghaiTechRGBD}, \ref{tab:miou_micc}, \ref{tab:miou_venice}. However, we still clearly outperform the baseline models. This shows that our approach does well even in the absence of a ground-truth depth map and is therefore applicable in a wide range of scenarios. In the supplementary material, we will show that our defense mechanism remains robust even when this pre-trained deep model is exposed to the attacker.

\subsection{Tampering with the Ground-Truth Depth Map}
\label{sec:robust}

The results of Sections~\ref{sec:compare} and~\ref{sec:pretrain}  were obtained under the assumption that the reference depth map is safe from attack. In some scenarios, this might not be the case and the attacker might be able to tamper with the depth map. Fortunately, even if this were the case, the attack would still be detectable as follows. Given $N$ training depth maps $\{\z_{1},\z_{2},...,\z_{N}\}$ recorded by a fixed depth sensor, we can record the min and max depth values for each pixel $j$ as
\begin{eqnarray}
    z_{min}^{j} &=&  \min(z_{1}^{j},z_{2}^{j},...,z_{N}^{j})\;, \\ \nonumber
    z_{max}^{j}  &=& \max(z_{1}^{j},z_{2}^{j},...,z_{N}^{j})\;.
\end{eqnarray}
Given $M$ reference depth maps  $\{\dot{\z}_{1},\dot{\z}_{2},...,\dot{\z}_{M}\}$ that are exposed and can be tampered with, the tampered depth value at pixel $j$ of the reference depth map $\dot{\z}_{k}$ can be written as
\begin{eqnarray}
    t_{k}^{j} = \dot{z}_{k}^{j}+\beta \mu_z^{j} \; .
\end{eqnarray}
where $\mu_z^{j}$  is the mean depth value in the reference depth maps in the test dataset, and $\beta$ is a scalar that represents the perturbation strength of a potential attack. We take the perturbation $\beta \mu_z^{j}$ to be a function of $\beta$ because the reference depth map is not an input to our network. If the attacker were able to choose an appropriate $\beta$ for each pixel of the reference depth map, the tampering indicator of Eq.~\ref{eq:indicator} could be compromised. Fortunately, such an attack is very likely to be detected using the following simple but effective approach. If $t_{k}^{j} < z_{min}^{j}$ or $t_{k}^{j} > z_{max}^{j}$, we label pixel $j$ as potentially tampered with. As shown  in Fig.~\ref{fig:depth_detection}, the pixel-wise detection accuracy is over 90\% even for extremely small perturbations with $\beta=0.01$ and quickly increases from there. This makes the attacker's task difficult indeed.

\subsection{Sensitivity Analysis}
\label{sec:sensitivity}


\begin{table}
  \centering
  \scalebox{0.8}{
  \rowcolors{2}{white}{gray!10}
  \begin{tabular}{lcccccccccccc}
    \toprule
      Dataset& \multicolumn{4}{c}{ShanghaiTechRGBD} & \multicolumn{4}{c}{MICC} & \multicolumn{4}{c}{Venice}\\
      Indicator value & $1\% $ & $3\% $ & $5\% $ & $10\% $ & $1\% $ & $3\% $ & $5\% $ & $10\% $ & $1\% $ & $3\% $ & $5\% $ & $10\% $\\
    \midrule
    \FGSMU{1}  & 0.42 & 0.48 & {\bf 0.54} & 0.51 & 0.38 & 0.40 & {\bf 0.46} & 0.42  & 0.38 & 0.40 & {\bf 0.42} & 0.40 \\
    \FGSMT{1} & 0.45 & 0.49& {\bf 0.54} &  0.49 & 0.40 & 0.42 & {\bf 0.49} &  0.46 & 0.36 & 0.42& {\bf 0.49} &  0.40\\
    \FGSMU{19} & 0.41 & 0.47 & {\bf 0.58}&  0.50 & 0.37 & 0.41& {\bf 0.49}&  0.43 & 0.35 & 0.38 & 0.36&  {\bf 0.45}\\
    \FGSMT{19} & 0.44 & 0.52 & {\bf 0.57}&  0.50 & 0.37 & 0.40 & {\bf 0.49}&  0.42 & 0.36 & {\bf 0.43} & 0.38&  0.33\\
    \FGSMUE{1} & 0.39&  0.43& {\bf 0.52}&  0.48 & 0.35&  0.38& 0.41&  {\bf 0.44} & 0.38&  {\bf 0.46}& 0.40&  0.38\\
    \FGSMTE{1}& 0.42& 0.44 & {\bf 0.51}& 0.46 & 0.33& 0.38& 0.41& {\bf 0.44} & 0.40& 0.42 & {\bf 0.48}& 0.41\\
    \FGSMUE{19} & 0.36 & 0.42 & {\bf 0.45} & 0.40 & 0.36 & {\bf 0.44} & 0.40 & 0.38 & 0.35 & 0.36 & 0.38 & {\bf 0.44}\\
    \FGSMTE{19} & 0.40& 0.42& {\bf 0.47} & 0.44 & 0.39& {\bf 0.46}& 0.40 & 0.38 & 0.38& {\bf 0.45}&  0.41 & 0.38\\
    \bottomrule
    \end{tabular}}
  \caption{{\bf Pixel-wise adversarial detection on different datasets for different indicator values.}}
  \label{tab:indicator_all}
\end{table}

\begin{table*}
  \centering
  \scalebox{0.6}{
    \rowcolors{2}{white}{gray!10}
    \begin{tabular}{lcccccccccccccccc}
      \toprule
   Dataset &\multicolumn{4}{c}{\FGSMUE{1}}& \multicolumn{4}{c}{\FGSMTE{1}}& \multicolumn{4}{c}{\FGSMUE{19}} & \multicolumn{4}{c}{\FGSMTE{19}}\\
  $\lambda$  & $mIoU$ & $DMAE$&  $RMSE$ & $ZMAE$ & $mIoU$  & $DMAE$&  $RMSE$ & $ZMAE$ & $mIoU$ & $DMAE$&  $RMSE$ & $ZMAE$ & $mIoU$ & $DMAE$&  $RMSE$ & $ZMAE$ \\
  \midrule
  0.01 & {\bf 0.52}& 58.14 & 68.34 & 0.11 & {\bf 0.51} &53.64 & 63.43 & 0.11 &{\bf 0.45} & 63.81 & 74.30  & 0.10 & {\bf 0.47}&52.89 & 62.43 & 0.10 \\
  1.0  & 0.41 & 36.62 & 42.51 & 0.09 &0.38& 35.73  &41.44 &  0.09 & 0.36 & 36.72 & 40.14 & 0.08 & 0.36 &  33.73& 38.12 & 0.08 \\
  100.0 & 0.36 & {\bf 18.73} & {\bf 22.31} & {\bf 0.08} &0.33 & {\bf 15.59} & {\bf 23.32} & {\bf 0.07}& 0.30 & {\bf 14.83} & {\bf 19.11} & {\bf 0.06}& 0.31&{\bf 17.62}& {\bf 20.68} & {\bf 0.07}\\
  \bottomrule
  \end{tabular}
  }
  \caption{{\bf Detection accuracy and error rates for different $\lambda$ values on ShanghaiTechRGBD.}}
  \label{tab:whitebox_error}
\end{table*}

We now quantify the influence of the three main hyper-parameters introduced in Section~\ref{sec:attacks} that control the intensity of the attacks.

\parag{Perturbation value.} We change the value of $\epsilon$ in Eq.~\ref{eq:FGSMu} and Eq.~\ref{eq:FGSMu1} from 1.0 to 35.0 for all attacks and plot the resulting mIoU in Fig.~\ref{fig:strength}.  Our model can detect very weak attacks with $\epsilon$ down to 1.0 and its performance quickly increases for larger values. In the supplementary material, we will exhibit the monotonous relationship between $\epsilon$ and the people density estimation error. When $\epsilon=1.0$, there is already a small perturbation of the density estimates---around 6 in $DMAE$ for {\bf ShanghaiTechRGBD}---that then become much larger as $\epsilon$ increases. The number of iterations $n$ is set to $min(\epsilon+4,1.25\epsilon)$ as recommended in earlier work~\cite{Kurakin16}.

\parag{Threshold value.} In Table~\ref{tab:indicator_all}, we report mIoU values on each dataset as a function of the threshold we use to classify a pixel as tampered with or not, depending on the ratio of Eq.~\ref{eq:indicator}. 5\% gives the best answer across all attacks.

\parag{Strength of Exposed Attacks.}

To check the robustness of our model against exposed attacks, we evaluate different $\lambda$ values in the loss term of Eq.~\ref{eq:FGSMue}, whose role is to keep the depth estimate as steady as possible in {\bf ShanghaiTechRGBD}. We tested our approach for values of $\lambda$ ranging from 0.01 to 100.0 and report the detection accuracy results along with the crowd counting error and depth error in Table~\ref{tab:whitebox_error}. For larger values $\lambda$, both the crowd density error and the detection rate drop.
In other words, increasing $\lambda$ makes the attack harder to detect but also weaker. We show the same trend in the other datasets in the supplementary material.


\section{Conclusion and Future Perspectives}

In this paper, we have shown that estimating density and depth jointly in a two-stream network could be leveraged to detect adversarial attacks against crowd-counting models at pixel level. Our experiments have demonstrated this to be the case even when the attacker knows our detection strategy, or has access to the reference depth map. In essence, our approach is an instance of a broader idea: One can leverage multi-task learning to detect adversarial attacks. In the future, we will therefore study the use of this approach for other tasks, such as depth estimation, optical flow estimation, and semantic segmentation.

\clearpage
%
%
\bibliographystyle{splncs04}
\bibliography{string,vision,learning}

\begin{thebibliography}{10}
\providecommand{\url}[1]{\texttt{#1}}
\providecommand{\urlprefix}{URL }
\providecommand{\doi}[1]{https://doi.org/#1}

\bibitem{Bhagoji17a}
Bhagoji, A., Cullina, D., Mittal, P.: {Dimensionality reduction as a defense
  against evasion attacks on machine learning classifiers}. In: arXiv preprint
  arXiv:1704.02654 (2017)

\bibitem{Bondi14a}
Bondi, E., Seidenari, L., Bagdanov, A., Bimbo, A.: {Real-time people counting
  from depth imagery of crowded environments}. International Conference on
  Advanced Video and Signal Based Surveillance  (2014)

\bibitem{Brostow06}
Brostow, G.J., Cipolla, R.: {Unsupervised Bayesian Detection of Independent
  Motion in Crowds}. In: Conference on Computer Vision and Pattern Recognition.
  pp. 594--601 (2006)

\bibitem{Cao18}
Cao, X., Wang, Z., Zhao, Y., Su, F.: {Scale Aggregation Network for Accurate
  and Efficient Crowd Counting}. In: European Conference on Computer Vision
  (2018)

\bibitem{Carlini17b}
Carlini, N., Wagner, D.: {Adversarial Examples Are Not Easily Detected:
  Bypassing Ten Detection Methods}. In: ACM Workshop on Artificial Intelligence
  and Security (2017)

\bibitem{Carlini17}
Carlini, N., Wagner, D.: {Towards Evaluating the Robustness of Neural
  Networks}. In: IEEE Symposium on Security and Privacy. pp. 39--57 (2017)

\bibitem{Chan08}
Chan, A., Liang, Z., Vasconcelos, N.: {Privacy Preserving Crowd Monitoring:
  Counting People Without People Models or Tracking}. In: Conference on
  Computer Vision and Pattern Recognition (2008)

\bibitem{Chan09}
Chan, A., Vasconcelos, N.: {Bayesian Poisson Regression for Crowd Counting}.
  In: International Conference on Computer Vision. pp. 545--551 (2009)

\bibitem{Chen12f}
Chen, K., Loy, C., Gong, S., Xiang, T.: {Feature Mining for Localised Crowd
  Counting}. In: British Machine Vision Conference. p.~3 (2012)

\bibitem{Cheng19a}
Cheng, Z., Li, J., Dai, Q., Wu, X., Hauptmann, A.G.: {Learning Spatial
  Awareness to Improve Crowd Counting}. In: International Conference on
  Computer Vision (2019)

\bibitem{Feinman17a}
Feinman, R., Curtin, R., Shintre, S., Gardner, A.: {Detecting Adversarial
  Samples from Artifacts}. In: preprint arXiv:1703.00410 (2017)

\bibitem{Fiaschi12}
Fiaschi, L., Koethe, U., Nair, R., Hamprecht, F.: {Learning to Count with
  Regression Forest and Structured Labels}. In: International Conference on
  Pattern Recognition. pp. 2685--2688 (2012)

\bibitem{Gal16}
Gal, Y., Ghahramani, Z.: {Dropout as a Bayesian Approximation: Representing
  Model Uncertainty in Deep Learning}. In: International Conference on Machine
  Learning. pp. 1050--1059 (2016)

\bibitem{Gong17a}
Gong, Z., Wang, W., Ku, W.: {Adversarial and clean data are not twins}. In:
  arXiv preprint arXiv:1704.04960 (2017)

\bibitem{Goodfellow2014a}
Goodfellow, I.J., Shlens, J., Szegedy, C.: {Explaining and Harnessing
  Adversarial Examples}. International Conference on Learning Representations
  (2015)

\bibitem{Grosse17}
Grosse, K., Manoharan, P., Papernot, N., Backes, M., McDaniel, P.: {On the
  (Statistical) Detection of Adversarial Examples}. In: arXiv preprint
  arXiv:1702.06280 (2017)

\bibitem{Metzen17}
Hendrik~Metzen, J., Genewein, T., Fischer, V., Bischoff, B.: {On Detecting
  Adversarial Perturbations}. International Conference on Learning
  Representations  (2017)

\bibitem{Idrees13}
Idrees, H., Saleemi, I., Seibert, C., Shah, M.: {Multi-Source Multi-Scale
  Counting in Extremely Dense Crowd Images}. In: Conference on Computer Vision
  and Pattern Recognition. pp. 2547--2554 (2013)

\bibitem{Idrees18}
Idrees, H., Tayyab, M., Athrey, K., Zhang, D., Al-maadeed, S., Rajpoot, N.,
  Shah, M.: {Composition Loss for Counting, Density Map Estimation and
  Localization in Dense Crowds}. In: European Conference on Computer Vision
  (2018)

\bibitem{Jiang19a}
Jiang, X., Xiao, Z., Zhang, B., Zhen, X.: {Crowd Counting and Density
  Estimation by Trellis Encoder-Decoder Networks}. In: Conference on Computer
  Vision and Pattern Recognition (2019)

\bibitem{Kendall17}
Kendall, A., Gal, Y.: {What Uncertainties Do We Need in Bayesian Deep Learning
  for Computer Vision?} In: Advances in Neural Information Processing Systems
  (2017)

\bibitem{Kurakin16}
Kurakin, A., Goodfellow, I., Bengio, S.: {Adversarial Machine Learning at
  Scale}. International Conference on Learning Representations  (2017)

\bibitem{Lakshminarayanan17}
Lakshminarayanan, B., Pritzel, A., Blundell, C.: {Simple and Scalable
  Predictive Uncertainty Estimation Using Deep Ensembles}. In: Advances in
  Neural Information Processing Systems (2017)

\bibitem{Lempitsky10}
Lempitsky, V., Zisserman, A.: {Learning to Count Objects in Images}. In:
  Advances in Neural Information Processing Systems (2010)

\bibitem{Li17b}
Li, X., Li, F.: {Adversarial Examples Detection in Deep Networks with
  Convolutional Filter Statistics}. In: International Conference on Computer
  Vision (2017)

\bibitem{Li18f}
Li, Y., Zhang, X., Chen, D.: {CSRNet: Dilated Convolutional Neural Networks for
  Understanding the Highly Congested Scenes}. In: Conference on Computer Vision
  and Pattern Recognition (2018)

\bibitem{Lian19a}
Lian, D., Li, J., Zheng, J., Luo, W., Gao, S.: {Density Map Regression Guided
  Detection Network for RGB-D Crowd Counting and Localization}. In: Conference
  on Computer Vision and Pattern Recognition (2019)

\bibitem{Lin10}
Lin, Z., Davis, L.: {Shape-Based Human Detection and Segmentation via
  Hierarchical Part-Template Matching}. IEEE Transactions on Pattern Analysis
  and Machine Intelligence  \textbf{32}(4),  604--618 (2010)

\bibitem{Lis19}
Lis, K., Nakka, K., Salzmann, M., Fua, P.: {Detecting the Unexpected via Image
  Resynthesis}. In: International Conference on Computer Vision (2019)

\bibitem{Liu19d}
Liu, C., Weng, X., Mu, Y.: {Recurrent Attentive Zooming for Joint Crowd
  Counting and Precise Localization}. Conference on Computer Vision and Pattern
  Recognition  (2019)

\bibitem{Liu18a}
Liu, J., Gao, C., Meng, D., Hauptmann1, A.: {Decidenet: Counting Varying
  Density Crowds through Attention Guided Detection and Density Estimation}.
  In: Conference on Computer Vision and Pattern Recognition (2018)

\bibitem{Liu19f}
Liu, L., Qiu, Z., Li, G., Liu, S., Ouyang, W., Lin, L.: {Crowd Counting with
  Deep Structured Scale Integration Network}. International Conference on
  Computer Vision  (2019)

\bibitem{Liu18c}
Liu, L., Wang, H., Li, G., Ouyang, W., Lin, L.: {Crowd Counting Using Deep
  Recurrent Spatial-Aware Network}. In: International Joint Conference on
  Artificial Intelligence (2018)

\bibitem{Liu19c}
Liu, N., Long, Y., Zou, C., Niu, Q., Pan, L., Wu, H.: {ADCrowdNet: An
  Attention-Injective Deformable Convolutional Network for Crowd
  Understanding}. Conference on Computer Vision and Pattern Recognition  (2019)

\bibitem{Liu19b}
Liu, W., Lis, K., Salzmann, M., Fua, P.: {Geometric and Physical Constraints
  for Drone-Based Head Plane Crowd Density Estimation}. International
  Conference on Intelligent Robots and Systems  (2019)

\bibitem{Liu19a}
Liu, W., Salzmann, M., Fua, P.: {Context-Aware Crowd Counting}. In: Conference
  on Computer Vision and Pattern Recognition (2019)

\bibitem{Liu18b}
Liu, X., Weijer, J., Bagdanov, A.: {Leveraging Unlabeled Data for Crowd
  Counting by Learning to Rank}. In: Conference on Computer Vision and Pattern
  Recognition (2018)

\bibitem{Liu19e}
Liu, Y., Shi, M., Zhao, Q., Wang, X.: {Point in, Box out: Beyond Counting
  Persons in Crowds}. Conference on Computer Vision and Pattern Recognition
  (2019)

\bibitem{Ma19a}
Ma, Z., Wei, X., Hong, X., Gong, Y.: {Bayesian Loss for Crowd Count Estimation
  with Point Supervision}. International Conference on Computer Vision  (2019)

\bibitem{Madry18}
Madry, A., Makelov, A., Schmidt, L., Tsipras, D., Vladu, A.: {Towards deep
  learning models resistant to adversarial attacks}. In: International
  Conference on Learning Representations (2018)

\bibitem{Moosavi16}
Moosavi-Dezfooli, S.M., Fawzi, A., Frossard, P.: {Deepfool: A Simple and
  Accurate Method to Fool Deep Neural Networks}. In: Conference on Computer
  Vision and Pattern Recognition. pp. 2574--2582 (2016)

\bibitem{Moosavi17}
Moosavi-Dezfooli, S., Fawzi, A., Fawzi, O., Frossard, P.: {Universal
  adversarial perturbations}. In: Conference on Computer Vision and Pattern
  Recognition (2017)

\bibitem{Onoro16}
Onoro-Rubio, D., L{\'o}pez-Sastre, R.: {Towards Perspective-Free Object
  Counting with Deep Learning}. In: European Conference on Computer Vision. pp.
  615--629 (2016)

\bibitem{Rabaud06}
Rabaud, V., Belongie, S.: {Counting Crowded Moving Objects}. In: Conference on
  Computer Vision and Pattern Recognition. pp. 705--711 (2006)

\bibitem{Ranjan19a}
Ranjan, A., Janai, J., Geiger, A., Black, M.J.: {Attacking Optical Flow}. In:
  International Conference on Computer Vision (2019)

\bibitem{Ranjan18}
Ranjan, V., Le, H., Hoai, M.: {Iterative Crowd Counting}. In: European
  Conference on Computer Vision (2018)

\bibitem{Sam18}
Sam, D., Sajjan, N., Babu, R., Srinivasan, M.: {Divide and Grow: Capturing Huge
  Diversity in Crowd Images with Incrementally Growing CNN}. In: Conference on
  Computer Vision and Pattern Recognition (2018)

\bibitem{Sam17}
Sam, D., Surya, S., Babu, R.: {Switching Convolutional Neural Network for Crowd
  Counting}. In: Conference on Computer Vision and Pattern Recognition. p.~6
  (2017)

\bibitem{Shen18}
Shen, Z., Xu, Y., Ni, B., Wang, M., Hu, J., Yang, X.: {Crowd Counting via
  Adversarial Cross-Scale Consistency Pursuit}. In: Conference on Computer
  Vision and Pattern Recognition (2018)

\bibitem{Shi19a}
Shi, M., Yang, Z., Xu, C., Chen, Q.: {Revisiting Perspective Information for
  Efficient Crowd Counting}. In: Conference on Computer Vision and Pattern
  Recognition (2019)

\bibitem{Shi19b}
Shi, Z., Mettes, P., Snoek, C.G.M.: {Counting with Focus for Free}. In:
  International Conference on Computer Vision (2019)

\bibitem{Shi18}
Shi, Z., Zhang, L., Liu, Y., Cao, X.: {Crowd Counting with Deep Negative
  Correlation Learning}. In: Conference on Computer Vision and Pattern
  Recognition (2018)

\bibitem{Sindagi17}
Sindagi, V., Patel, V.: {Generating High-Quality Crowd Density Maps Using
  Contextual Pyramid CNNs}. In: International Conference on Computer Vision.
  pp. 1879--1888 (2017)

\bibitem{Sindagi19a}
Sindagi, V., Patel, V.: {Multi-Level Bottom-Top and Top-Bottom Feature Fusion
  for Crowd Counting}. In: International Conference on Computer Vision (2019)

\bibitem{Szegedy13}
Szegedy, C., Zaremba, W., Sutskever, I., Bruna, J., Erhan, D., Goodfellow, I.,
  Fergus, R.: {Intriguing Properties of Neural Networks}. arXiv Preprint
  (2013)

\bibitem{Wan19b}
Wan, J., Chan, A.B.: {Adaptive Density Map Generation for Crowd Counting}. In:
  International Conference on Computer Vision (2019)

\bibitem{Wan19a}
Wan, J., Luo, W., Wu, B., Chan, A.B., Liu, W.: {Residual Regression with
  Semantic Prior for Crowd Counting}. In: Conference on Computer Vision and
  Pattern Recognition (2019)

\bibitem{Wang19a}
Wang, Q., Gao, J., Lin, W., Yuan, Y.: {Learning from Synthetic Data for Crowd
  Counting in the Wild}. In: Conference on Computer Vision and Pattern
  Recognition (2019)

\bibitem{Wang11a}
Wang, X., Wang, B., Zhang, L.: {Airport Detection in Remote Sensing Images
  Based on Visual Attention}. In: International Conference on Neural
  Information Processing (2011)

\bibitem{Wu05}
Wu, B., Nevatia, R.: {Detection of Multiple, Partially Occluded Humans in a
  Single Image by Bayesian Combination of Edgelet Part Detectors}. In:
  International Conference on Computer Vision (2005)

\bibitem{Xiao18}
Xiao, C., Deng, R., Li, B., Yu, F., Liu, M., Song, D.: Characterizing
  adversarial examples based on spatial consistency information for semantic
  segmentation. In: European Conference on Computer Vision. pp. 217--234 (2018)

\bibitem{Xiong17}
Xiong, F., Shi, X., Yeung, D.: {Spatiotemporal Modeling for Crowd Counting in
  Videos}. In: International Conference on Computer Vision. pp. 5161--5169
  (2017)

\bibitem{Xiong19a}
Xiong, H., Lu, H., Liu, C., Liu, L., Cao, Z., Shen, C.: {From Open Set to
  Closed Set: Counting Objects by Spatial Divide-and-Conquer}. In:
  International Conference on Computer Vision (2019)

\bibitem{Xu19a}
Xu, C., Qiu, K., Fu, J., Bai, S., Xu, Y., Bai, X.: {Learn to Scale: Generating
  Multipolar Normalized Density Maps for Crowd Counting}. In: International
  Conference on Computer Vision (2019)

\bibitem{Yan19a}
Yan, Z., Yuan, Y., Zuo, W., Tan, X., Wang, Y., Wen, S., Ding, E.:
  {Perspective-Guided Convolution Networks for Crowd Counting}. In:
  International Conference on Computer Vision (2019)

\bibitem{Yin19a}
Yin, W., Liu, Y., Shen, C., Yan, Y.: {Enforcing geometric constraints of
  virtual normal for depth prediction}. In: International Conference on
  Computer Vision (2019)

\bibitem{Zhang19b}
Zhang, A., Shen, J., Xiao, Z., Zhu, F., Zhen, X., Cao, X., Shao, L.:
  {Relational Attention Network for Crowd Counting}. In: International
  Conference on Computer Vision (2019)

\bibitem{Zhang19c}
Zhang, A., Yue, L., Shen, J., Zhu, F., Zhen, X., Cao, X., Shao, L.:
  {Attentional Neural Fields for Crowd Counting}. In: International Conference
  on Computer Vision (2019)

\bibitem{Zhang15c}
Zhang, C., Li, H., Wang, X., Yang, X.: {Cross-Scene Crowd Counting via Deep
  Convolutional Neural Networks}. In: Conference on Computer Vision and Pattern
  Recognition. pp. 833--841 (2015)

\bibitem{Zhang19a}
Zhang, Q., Chan, A.B.: {Wide-Area Crowd Counting via Ground-Plane Density Maps
  and Multi-View Fusion CNNs}. In: Conference on Computer Vision and Pattern
  Recognition (2019)

\bibitem{Zhang16s}
Zhang, Y., Zhou, D., Chen, S., Gao, S., Ma, Y.: {Single-Image Crowd Counting
  via Multi-Column Convolutional Neural Network}. In: Conference on Computer
  Vision and Pattern Recognition. pp. 589--597 (2016)

\bibitem{Zhao19a}
Zhao, M., Zhang, J., Zhang, C., Zhang, W.: {Leveraging Heterogeneous Auxiliary
  Tasks to Assist Crowd Counting}. In: Conference on Computer Vision and
  Pattern Recognition (2019)

\end{thebibliography}
\end{document}